\documentclass[runningheads]{llncs}

\usepackage{eccv}

\usepackage{eccvabbrv}

\usepackage{graphicx}
\usepackage{booktabs}
\usepackage{bm}
\usepackage{pifont}

\newcommand{\actionscore}{\ensuremath{S_{\mathrm{cons}}}}
\newcommand{\tempscore}{\ensuremath{S_{\mathrm{temp}}}}
\newcommand{\actionname}{\emph{Action Consistency}}
\newcommand{\tempname}{\emph{Temporal Coherence}}
\newcommand{\cmark}{\textcolor{gray}{\ding{51}}} 
\newcommand{\xmark}{\textcolor{red}{\ding{55}}}      
\usepackage{fontawesome}
\usepackage{booktabs}
\usepackage{makecell}
\usepackage{float}      
\usepackage{placeins}   
\usepackage{tablefootnote}
\usepackage{listings}
\usepackage{fancyvrb}
\usepackage{fvextra}
\usepackage[font=scriptsize]{caption}
\usepackage{fvextra}

\usepackage[accsupp]{axessibility}  

\usepackage{hyperref}

\usepackage{orcidlink}

\begin{document}

\title{Generative Action Tell-Tales: Assessing Human Motion in Synthesized Videos} 

\titlerunning{Generative Action Tell-Tales}




\author{
Xavier Thomas\inst{1} \and
Youngsun Lim\inst{1} \and
Ananya Srinivasan\inst{2} \and
Audrey Zheng\inst{3} \and
Deepti Ghadiyaram\inst{1,4}
}

\authorrunning{X. Thomas et al.}

\institute{
Boston University \and
Belmont High School \and
Canyon Crest Academy \and
Runway
}

\maketitle

\begin{abstract}
    Despite rapid advances in video generative models, robust metrics for evaluating visual and temporal correctness of complex human actions remain elusive. Critically, existing pure-vision encoders and Multimodal Large Language Models (MLLMs) are strongly appearance-biased, lack temporal understanding, and thus struggle to discern intricate motion dynamics and anatomical implausibilities in generated videos. We tackle this gap by introducing a novel evaluation metric derived from a learned latent space of real-world human actions. Our method first captures the nuances, constraints, and temporal smoothness of real-world motion by fusing appearance-agnostic human skeletal geometry features with appearance-based features. We posit that this combined feature space provides a robust representation of action plausibility. Given a generated video, our metric quantifies its action quality by measuring the distance between its underlying representations and this learned real-world action distribution. For rigorous validation, we develop a new multi-faceted benchmark specifically designed to probe temporally challenging aspects of human action fidelity. Through extensive experiments, we show that our metric achieves substantial improvement of more than $\mathbf{68\%}$ compared to existing state-of-the-art methods on our benchmark, performs competitively on established external benchmarks, and has a stronger correlation with human perception. Our in-depth analysis reveals critical limitations in current video generative models and establishes a new standard for advanced research in video generation. Code is available at \href{https://xthomasbu.github.io/video-gen-evals/}{https://xthomasbu.github.io/video-gen-evals/}
\end{abstract}

\section{Introduction}\label{sec:1}

\begin{figure}[tb]
\centering
\includegraphics[width=0.47\linewidth]{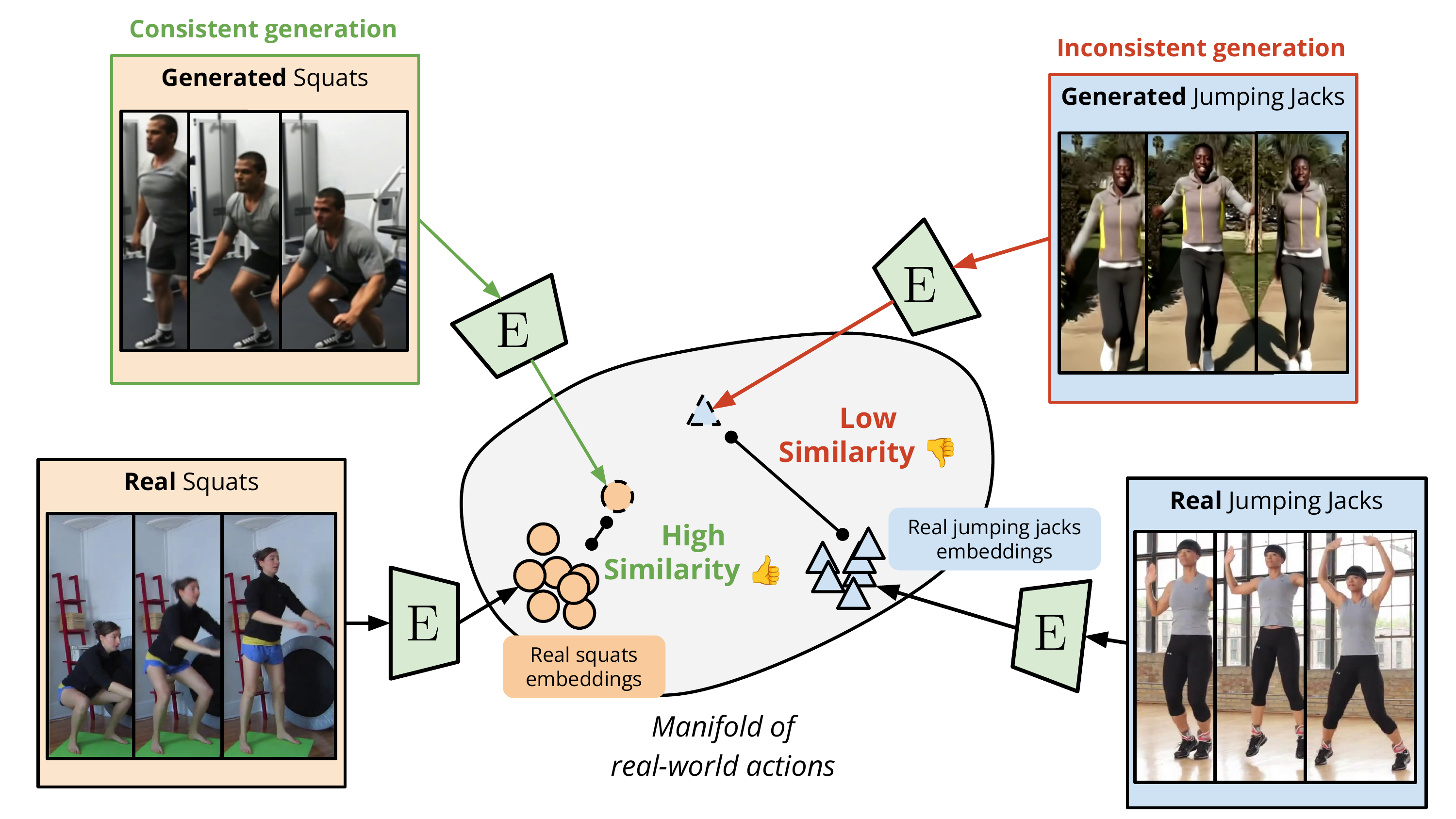}%
\hspace{0.01\linewidth}%
\vrule width 0.3pt%
\hspace{0.01\linewidth}%
\includegraphics[width=0.50\linewidth]{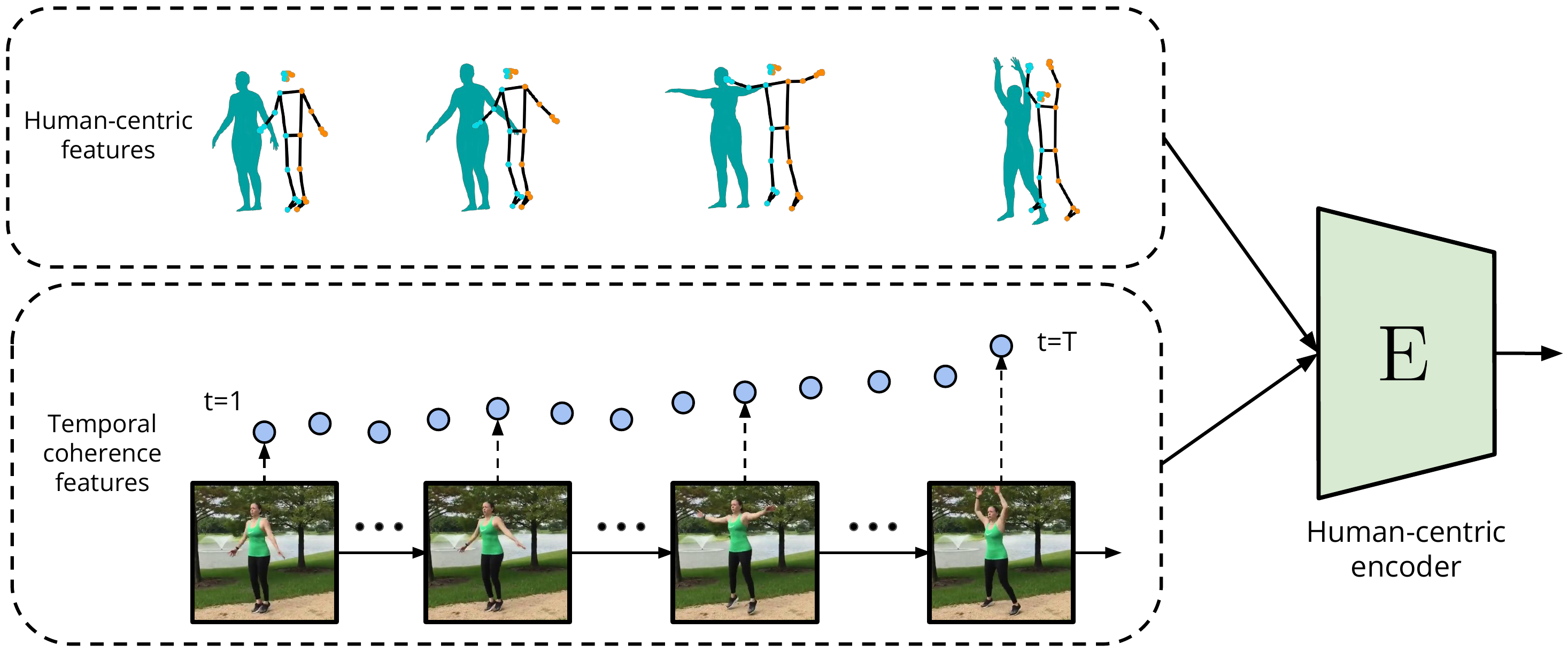}
\caption{What are the \textbf{telltale signs of a generative action video}? 
We answer this by learning a robust manifold based on appearance and anatomical coherence exhibited by humans performing actions across several real-world videos. 
This manifold serves as anchors against which we project the features of a generated video in question and assess its realism.}
\vspace{-1.0em}
\label{fig:teaser}
\end{figure}

How do humans learn the right way to perform an action, like \textit{walking} or \textit{making a toast}? Since infancy, we learn by implicitly observing and explicitly being taught by others, grasping the flow of events and the physical laws governing human motions and actions~\cite{gopnik}. This intuitive understanding allows humans to effortlessly recognize motion inconsistencies even in today’s highly photorealistic generated videos~\cite{physicsIQ,runway2024gen4, wan2025wan}.

This raises a critical question: \textit{Can we formalize our intuitive perception and design a framework to systematically evaluate the accuracy of human actions in generated videos?} This is a challenging task, as it requires solving a twofold challenge. First, the baseline problem of recognizing action correctness is ill-posed, even in real videos. Actions can be atomic~\cite{gu2018ava} (e.g., ``walking'') or procedural (e.g., ``making a toast'')~\cite{barker}, making automatic recognition inherently challenging. Second, for generated videos, this problem is magnified as we must move beyond detecting the mere \textbf{presence of an action} to also evaluating the \textbf{temporal coherence} of the body movements over time.

Current metrics for judging generated videos, such as pixel-level similarity~\cite{psnr, ssim}, perceptual quality~\cite{lpips, FVD}, or text-to-video prompt alignment~\cite{clipscore, evalcrafter}, as well as recent approaches that use MLLMs as judges~\cite{videoscore, videophy}, do not accurately capture the complex motion physics of human actions~\cite{tevet2022human, cho2024sora}.
We posit that this is because the underlying representations fundamentally lack awareness of subtle anatomical violations or temporal incoherence. A core contribution of our work is to bridge this gap by building a robust assessment tool that moves beyond superficial statistics and bakes in the critical awareness of physical and anatomical-consistency of human motion.

Our key idea is to learn a latent manifold of natural human motion. This manifold is built from semantic features that measure the consistency of human anatomy, motion physics, and visual appearance in a video. As illustrated in ~\cref{fig:teaser}, we learn this manifold from a large volume of diverse videos capturing a wide range of body geometries and performance styles (e.g., a tall woman walking vs. an older male with a walking stick), thereby encapsulating the crucial cues of natural action. To evaluate a new video, we project its embeddings into this manifold and systematically measure the deviations and discern telltale signs of poor action quality, correctness, and coherence.

While several benchmarks exist~\cite{vbench2, evalcrafter}, they fail to adequately probe for the fine-grained temporal correctness and coherence of human actions. We identify this as a critical limitation and create an open-source benchmark we call the ``\textbf{T}elltale \textbf{A}ction \textbf{G}eneration Bench'' (\textbf{TAG-Bench-v0}). We generate hundreds of videos from state-of-the-art open- and closed-source models and conduct a large-scale subjective study. We evaluate the videos on two key criteria: (a) whether the generated video captures the intended action, and (b) the temporal smoothness and anatomical plausibility of the perceived action.

Our extensive experiments on \textbf{TAG-Bench-v0} and other benchmarks~\cite{vbench2} yield several key findings. First, we highlight a critical gap in current evaluation: we show that \underline{all} state-of-the-art models and MLLMs struggle to correctly evaluate action correctness and temporal coherence. This demonstrates the sheer difficulty of the task. We further find that certain actions are challenging for all generative models, revealing broader limitations in current video synthesis capabilities. Second, we demonstrate that, unlike existing methods, our manifold's scores strongly align with human opinion, across both image-to-video (I2V)~\cite{zhang2023i2vgen} and text-to-video (T2V)~\cite{blattmann2023stable} generation settings, and across multiple models. In summary, our contributions are:
\begin{itemize}
    \item We design a learned latent space that encodes human body geometry and temporal coherence to evaluate action quality in generated videos.
    \item We design \textbf{TAG-Bench-v0}, a benchmark with human ratings focused on the correctness and temporal coherence of human actions in generated videos.
    \item We propose two metrics that align closely with human perception. They measure the consistency and the temporal plausibility of actions.
    \item We provide an in-depth analysis of the proposed latent action manifold, and validate the design choices that led to its robust performance.
\end{itemize}

\section{Related Work}\label{sec:2}

\noindent\textbf{Video generation models.} 
Video generation~\cite{runway2024gen4, hong2022cogvideo, kong2024hunyuanvideo, wan2025wan, openai2024sora} has rapidly advanced, producing increasingly photorealistic and temporally coherent content~\cite{xing2024survey, videodiffusion}. 
Beyond visual fidelity, recent work~\cite{wiedemer2025video} frames these models as emerging \emph{world models} capable of capturing complex real-world dynamics without explicit supervision. 
Despite such progress, current systems still often generate human motion that is kinematically implausible or semantically incorrect~\cite{tevet2022human, cho2024sora}. This highlights the need for metrics that specifically diagnose and quantify these failures.

\noindent\textbf{Distribution and reference based metrics.}
Metrics such as FVD~\cite{FVD} measure statistical similarity between real and generated videos in pretrained feature spaces, capturing coarse spatiotemporal alignment but missing fine-grained semantic or physical accuracy. 
Frame-level measures including PSNR~\cite{psnr}, SSIM~\cite{ssim}, and LPIPS~\cite{lpips} assess visual similarity to reference frames but ignore temporal coherence critical to video perception. 
Recent efforts like Physics-IQ~\cite{physicsIQ} measure spatiotemporal realism using real-world references, but do not focus on capturing human-body distortions or action-level correctness.

\noindent\textbf{Reference-free video metrics.} 
In many scenarios, ground-truth videos are unavailable, motivating the need for reference-free evaluation. 
CLIPScore~\cite{clipscore} measures frame-text similarity scores but only captures single-frame semantics, lacking motion or physical plausibility.
More recent methods employ MLLMs for richer video understanding, via zero-shot prompting or fine-tuning on human ratings.
For instance, VideoScore~\cite{videoscore} predicts fine-grained human ratings across dimensions such as visual quality and temporal consistency, while VideoPhy~\cite{videophy} assesses whether generated videos follow physical laws like gravity or buoyancy.
However, neither captures human-body distortions or action correctness, underscoring the need for our proposed metric.

\noindent\textbf{Benchmarking video models.} 
With advances in generative video, several benchmarks now evaluate videos across multiple dimensions. EvalCrafter~\cite{evalcrafter} provides a large-scale framework for assessing visual quality, motion quality, temporal consistency, and text–video alignment. VBench2.0~\cite{vbench2} extends this with finer-grained anatomy-related criteria. 
However, none explicitly assess whether human actions are executed plausibly over time. Our proposed TAG-Bench-v0 addresses this gap through targeted metrics and a curated evaluation set designed to quantify human action realism and physically plausible motion.

\section{Approach}\label{sec:3}
We posit the \emph{``realism''} of human actions in generated videos as the distance between real and generated samples within a learned representation space. In this space, real human actions cluster into a compact region of natural, physically plausible movements. We refer to this as the action manifold (\cref{fig:teaser}). Capturing this notion of realism requires accurately modeling the \textit{temporal intrinsics} of the human body, human-object interactions, and the sequence of atomic actions involved in performing a given action. 

This section is structured as follows: we detail the variety of human-centric features in \cref{sec:human_features} that serve as building blocks of the learned action mainfold, which we describe in \cref{sec:embedding_model}. Next, we define the distance metric we learn to assess realism in generated videos in \cref{sec:embedding_metrics}.

\subsection{Human-centric feature representations}
\label{sec:human_features}
We capture the complexity of human motion leveraging human-centric features that measure appearance, skeletal geometry, and motion dynamics detailed next.
\subsubsection{3D features.} \label{sec:3d_feats}
We employ Skinned Multi-Person Linear (SMPL)~\cite{loper2023smpl}, a standard 3D representation of the human body. Briefly, SMPL consists of a ``rest pose'', a 3D mesh representation of an average human body, that is deformed based on three parameters: pose ($\theta$), body shape ($\beta$), and global orientation ($go$). $\mathbf{\theta}$ represents the 3D rotations of skeletal joints~\cite{loper2023smpl}, capturing how the human body moves during a particular action. $\mathbf{\beta}$ captures shape-specific characteristics~\cite{loper2023smpl}, such as overall build (e.g., tall vs. short) and limb proportions. $go$ describes the global body rotation computed from the pelvis joint~\cite{smpl_faq}, capturing how the orientation of the person's entire body changes during an action (e.g., turning sideways). 
To infer these SMPL features from 2D video frames, we rely on human mesh recovery (HMR)~\cite{hmrKanazawa17} models.

\noindent \textbf{Motivation to use SMPL.} We believe that the detailed 3D features are crucial to capture the complex kinematics of a human body in action. SMPL features are invariant to appearance and scene context~\cite{loper2023smpl, kocabas2020vibe}, and focus specifically on the geometry of the human body. They serve as powerful ingredients to assess if generated humans follow the same physical dynamics as real-world humans. 

\subsubsection{2D features.}
\label{sec:2d_feats}
While 3D representations capture detailed joint rotations and body shapes, SMPL is trained solely on real human data~\cite{loper2023smpl}, constraining its parameters to remain within anatomically plausible body configurations~\cite{pavlakos2019expressive}. 
This may overlook anomalies such as elongated limbs or implausible joint configurations common in \emph{generated} videos. Thus, we also incorporate 2D joint keypoints~\cite{cao2017realtime} ($kp_{2D}$), which are void of such strong priors. These 2D cues complement the 3D features by revealing anatomical distortions that the SMPL representation might overlook.

\subsubsection{Visual appearance features.}
\label{sec:app_feats}
Although 3D and 2D features capture body structure, they are inherently and intentionally appearance-invariant~\cite{loper2023smpl, cao2017realtime, clip}. To complement them, we extract visual appearance features ($f_{vis}$) using pre-trained image-based backbones (e.g., ViT~\cite{dosovitskiy2020image}). These features capture cues such as clothing, color, and action-relevant objects, all of which influence how an action is visually perceived. 

\subsubsection{First-order temporal coherence.}
\label{sec:motion_feats}
We refer to the human-centric features described so far ($\mathbb{S} = \{$$\theta$, $\beta$, $go$, $kp_{2D}$, $f_{vis}$ $\}$) as \emph{static} features, as they capture the body's state at a particular point in time (i.e., a frame). However, human actions inherently involve body dynamics that temporally evolve in a coherent manner.
Consider a generated video of a person performing bar pull-ups where the person’s arms become unrealistically long over time.
We believe that this unnatural body morphing will strongly emerge when we compute the temporal derivative of the body shape feature and will serve as a critical signal. Motivated by this, we compute the first-order temporal derivatives of each static feature, yielding corresponding \emph{motion} features: ($\mathbb{U} = \{$$m_{\theta}$, $m_{\beta}$, $m_{go}$, $m_{kp_{2D}}$, $m_{f_{vis}} \}$). These features make the action manifold sensitive to generative artifacts such as sudden body shape changes, jitter, or implausible pose transitions.

By combining 3D and 2D skeletal and appearance features and their corresponding temporal derivatives, we obtain a human-centric representation that is anatomically grounded and captures the natural evolution of human motion. Yet, we acknowledge that HMR and pose features may be unreliable under occlusion.

\begin{figure}[t]
  \centering
  \includegraphics[width=0.95\linewidth, keepaspectratio]{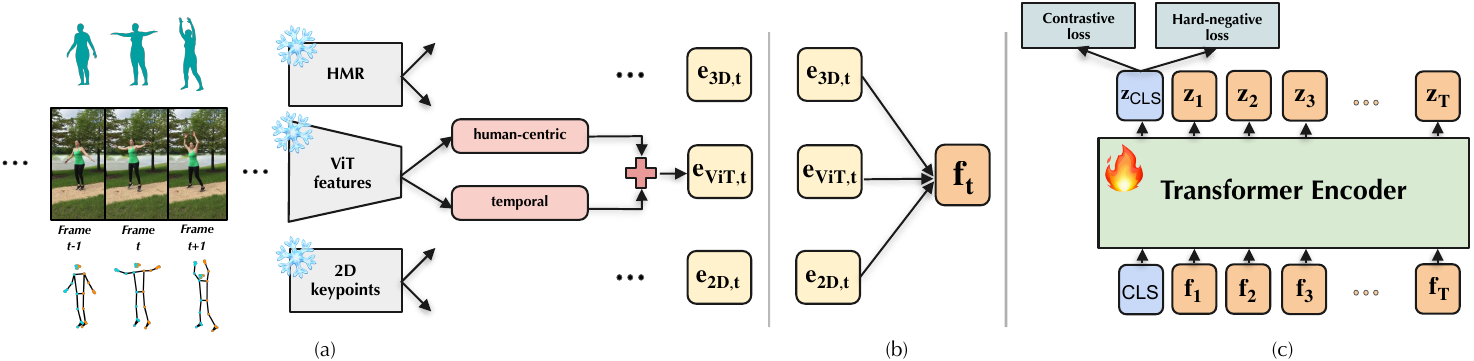}
  \vspace{-1.0em}
  \caption{ \textbf{Architectural overview of the encoder we train to learn the real-world action manifold.} We extract per-frame static human-centric and temporal motion features (Fig. (a)) (Sec.~\ref{sec:human_features}), and aggregate them, yielding one embedding for each frame (Fig. (b)) (Sec.~\ref{sec:model_training}). We prepend a $[\text{CLS}]$ token to the per-frame tokens and pass as input to a 4-layer transformer encoder (Fig. (c)) (Sec.~\ref{sec:model_training}). Our aim is to encourage the encoder to group diverse videos pertaining to a given action closer together. We also ensure that temporally incoherent videos lie farther apart.
  }
  \vspace{-2.0em}
  \label{fig:arch}
\end{figure}
\subsection{Learning a manifold of real-world actions}
\label{sec:embedding_model}
Our goal is to learn a compact human-centric \emph{latent representation} that captures the nuances of real-world actions. In this space, physically plausible actions occupy compact regions, while anatomically distorted or temporally inconsistent actions lie farther apart. To this end, we train an encoder to distinguish physically plausible motion from implausible human actions, as described next.

\subsubsection{Encoder architecture and training.}
\label{sec:model_training}

We represent each video as a sequence of fixed-length \emph{temporal windows}, each containing $T$ consecutive frames. This design allows the model to capture local, fine-grained motion of an action (e.g., a stride of a person running). For each frame $t \in \{1, \ldots, T\}$ in a window, we extract human-centric features from \cref{sec:human_features}, which serve as input to our model (\cref{fig:arch}(a)). The proposed model operates in three stages: (i) encode each input feature independently, (ii) fuse resulting per-input representations for each frame, and (iii) temporally aggregate these frame-level representations over the temporal window. We describe each stage in detail next.

\noindent \textbf{(i) Per-input encoding:} 
We take inspiration from~\cite{simonyan2014two} and define two pathways, one for \textit{static} ($\mathbb{S}$) and another for \textit{motion} ($\mathbb{U}$) features (\cref{sec:human_features}). To this end, let $\phi$ denote a $1$D temporal convolution block that aggregates features within a short temporal context. Each static and motion input feature ($s_{k,t} \in \mathbb{S}$ and $u_{k,t} \in \mathbb{U}$ respectively, where $k$ represents the distinct feature sources (i.e., pose, shape, keypoints, etc) in \cref{sec:human_features}) is processed using separate temporal convolution blocks, $\phi^k_{\text{static}}$ and $\phi^k_{\text{motion}}$. These per-input temporal convolutions help capture the local temporal evolution of each input feature (e.g., $\theta$).

Next, we combine the static and motion encodings for each frame $t$ via element-wise addition, similar to the additive fusion used in GENMO~\cite{li2025genmo}.
Thus, the motion pathway provides residual information to the  static pathway:

\begin{equation}
\label{eqn:state_motion}
\mathbf{e}_{k,t} = \phi^k_{\text{static}}(s_{k,t}) + \phi^k_{\text{motion}}(u_{k,t}),
\end{equation}
where $u_{k,t}$ is the temporal derivative of the corresponding static feature $s_{k,t}$, and $\mathbf{e}_{k,t} \in \mathbb{R}^d$, where $d$ is the dimension of the encoded input feature.  

\noindent\textbf{(ii) Per-frame feature fusion:} We next aggregate the encoded input features $\mathbf{e}_{k,t}$ to obtain a single frame-level representation using a learned attention mechanism. 
Specifically, for each frame $t$ and input $k$, the model assigns a scalar weight $\alpha_{k,t}$, capturing its relative importance for that frame. The fused representation $\mathbf{f}_t$ is then computed as a weighted sum of all features:
\begin{equation}
\mathbf{f}_t = \sum_k \alpha_{k,t}\,\mathbf{e}_{k,t}, \quad 
\alpha_{k,t} = \mathrm{softmax}_k\left(\frac{\mathbf{q}^\top \mathbf{W}_a \mathbf{e}_{k,t}}{\sqrt{d}}\right),
\end{equation}
The attention weights $\alpha_{k,t}$ are computed via a scaled dot-product attention~\cite{vaswani2017attention} between a learnable query vector $\mathbf{q} \in \mathbb{R}^{d}$ and a linear projection of each input feature using $\mathbf{W}_a \in \mathbb{R}^{d \times d}$. Softmax operation ensures that the weights are positive and sum to $1$. This attention mechanism is learned jointly with the model. 

\noindent\textbf{(iii) Temporal aggregation:} To aggregate all the fused frame representations $\{\mathbf{f}_t\}_{t=1}^{T}$ for a given temporal window, we prepend a learnable token denoted as $[\text{CLS}]$~\cite{timesformer}, and process the sequence with a Transformer encoder, which models long-range temporal dynamics (\cref{fig:arch}(c)). This results in a compact embedding $\mathbf{z}_{\text{CLS}}$ that captures the essential information over a temporal window. In addition, we extract the sequence of frame-level output embeddings $\{\mathbf{z}_t\}_{t=1}^{T}$ (\cref{fig:arch}(c)) to capture a finer-grained per-frame representation.

\subsubsection{Training objective.}
\label{sec:objective}
Our goal is to learn an effective latent space of real-world human actions. We achieve this via two complementary losses:

\noindent\textbf{(i) Learn action semantics:} We use a supervised contrastive loss  ($\mathcal{L}_{\text{supcon}}$)~\cite{animesh2023tunedcontrastivelearning} which encourages window-level embeddings ($\mathbf{z}_{\text{CLS}}$) from the same action class to cluster together in the latent space, while pushing embeddings from different action classes apart. This results in representations that are discriminative across actions, facilitating a robust understanding of what action is being performed.

\noindent\textbf{(ii) Enforcing temporal coherence:} In addition to distinguishing between actions, we also want our learned manifold to be temporally sensitive. For instance, a generated ``jumping jacks'' video may contain correct poses yet appear unrealistic if the person remains frozen mid-motion intermittently. To enforce this property in a self-supervised manner, we simulate temporally distorted variants of real videos by: (i) shuffling frames (breaking motion continuity), (ii) repeating the first frame across the window (simulating static frames), and (iii) reversing frame order (disrupting causal progression). We then introduce an additional loss ($\mathcal{L}_{\text{hard-negative}}$) that penalizes temporal inconsistency by pushing embeddings of these distorted windows away from their temporally coherent counterparts, teaching the model to recognize plausible motion dynamics.
Crucially, this objective goes beyond standard semantic negatives (i.e., different actions) and explicitly teaches the model to be sensitive to temporal structure. 

The overall training loss is a weighted sum between the two objectives:
\begin{equation} \label{eq:loss_total}
\mathcal{L} = \mathcal{L}_{\text{supcon}} + \lambda \mathcal{L}_{\text{hard-negative}},
\end{equation}
where $\lambda$ is a scalar weighting coefficient that balances the two loss terms. This combined objective encourages the encoder to capture both action semantics (\emph{what} action) and temporal coherence (\emph{how} the action is performed).

\subsection{Quantitative Metrics}
\label{sec:embedding_metrics}

From this learned embedding space, we derive quantitative measures to assess how closely a generated video aligns with the manifold of real human actions based on two key observations:
\begin{itemize}
    \item Real videos of the same action (e.g., ``jumping jacks'') form compact, \textbf{action consistent} clusters (\cref{fig:teaser}).
    \item Frame-level embeddings of a real video evolve smoothly over time, measuring \textbf{temporal coherence} (\cref{fig:teaser}).
\end{itemize}
Given a presumably poorly generated video, it should violate these two properties. To measure \textbf{action consistency ($\bm{\actionscore}$)}, 
we compute an action-specific centroid $\mathbf{c}_k$ by averaging the temporal window-level embeddings across all real videos of a given class $k$.
For a generated video of the same action, we similarly average its window-level $[\text{CLS}]$ embeddings to obtain $\mathbf{z}^{\text{gen}}_{\text{video}}$. A generated video is more action consistent if the distance between these two representations (${\actionscore}$ = $\left\lVert \mathbf{z}^{\text{gen}}_{\text{video}} - \mathbf{c}_k \right\rVert_2$) is small. Lower distance indicates closer alignment with real action distributions. To evaluate \textbf{temporal coherence ($\bm{\tempscore}$)}, we measure the smoothness of frame trajectories within the learned embedding space. Given per-frame embeddings ($\mathbf{z}_t$) of a window, we define temporal coherence as
\begin{equation}
\label{eq:smoothness_score}
{\tempscore} = \frac{1}{T-1} \sum_{t=1}^{T-1} \lVert \mathbf{z}_{t+1} - \mathbf{z}_t \rVert_2.
\end{equation}
The final score for a video is obtained by averaging temporal coherence scores across all windows. Lower scores correspond to gradual, physically consistent transitions, while higher scores indicate abrupt or implausible temporal changes. While action consistency measures if a generated action is semantically correct, temporal coherence measures how temporally smooth the generation is. In Sec.~\ref{sec:baselines}, we show that these metrics strongly correlate with human perception and serve as reliable indicators for evaluating generative video models.

\section{Telltale Action Generation (TAG)-Bench} \label{sec:human_eval}

As mentioned earlier, existing benchmarks~\cite{vbench2, videoscore2} do not probe for the finer-grained temporal correctness and coherence of human actions. To bridge this limitation, we build an open-source benchmark, ``\textbf{T}elltale \textbf{A}ction \textbf{G}eneration (TAG)-Bench-v0''. TAG-Bench-v0 comprises videos generated from several open- and closed-source I2V generation models associated with rich human opinion scores. We describe the data curation and human annotation process below.

\noindent{\textbf{Dataset construction:}} We select $10$ action classes from UCF-101~\cite{soomro2012ucf101} that (1) feature a single visible person, (2) depict the full human body, and (3) involve diverse whole-body movement dynamics: \textit{BodyWeightSquats, HulaHoop, JumpingJack, PullUps, PushUps, Shotput, SoccerJuggling, TennisSwing, ThrowDiscus,} and \textit{WallPushups}. By covering both localized (e.g., push-ups) and complex full-body dynamics (e.g., tennis swings), the selected classes allow for a meaningful evaluation of human motion synthesis quality. For each action, we sample $6$ videos at random and extract their first frame. This frame serves as the input to I2V generation models -- Wan2.1~\cite{wan2025wan}, Wan2.2~\cite{wan2025wan}, Hunyuan~\cite{kong2024hunyuanvideo}, Opensora~\cite{zheng2024open}, and Runway Gen-4~\cite{runway2024gen4} -- along with the prompt ``A person is doing \{action\}.'' This yields $300$ generated videos (more in Appendix).
We focus on I2V to ensure all models begin from the same visual input (i.e., the initial frame), allowing us to isolate differences in motion generation capabilities without confounding factors like variations in scene layout or the person’s appearance. 

\noindent \textbf{Video annotation setup:} We recruit $246$ participants through Amazon Mechanical Turk~\cite{mturk}. Each participant rated the videos on a $1$--$10$ scale along two axes: (1) \textbf{\actionname}, i.e., how accurately the generated video depicts the intended action mentioned in the prompt and (2) \textbf{\tempname}, i.e., how physically plausible and temporally smooth the motion appears in the generated video.  After subject rejection, the average inter-rater correlation reached $0.716$ for {\actionname} and $0.710$ for {\tempname}. Additional details in Appendix.

\section{Experiments}\label{sec:5}

\begin{table}[t]
\small
\centering
\setlength{\tabcolsep}{4pt}
\renewcommand{\arraystretch}{0.95}
\resizebox{0.85\linewidth}{!}{
\begin{tabular}{l|cc}
\toprule
\textbf{Method} &
\makecell[c]{\textbf{Corr. with Action}\\\textbf{Consistency} \textcolor{ForestGreen}{$\uparrow$}} &
\makecell[c]{\textbf{Corr. with Temporal}\\\textbf{Coherence} \textcolor{ForestGreen}{$\uparrow$}} \\
\midrule
Random & -0.07 & -0.11 \\
\midrule
\multicolumn{3}{c}{\textbf{Feature-based automatic metrics (Top-3)}} \\
\midrule
VideoMAE(UCF101)-classification        & 0.18 & 0.17 \\
DINO-sim \cite{dino}                     & 0.08 & 0.21 \\
CLIP-sim \cite{clip}                    & 0.03 & 0.16 \\
\midrule
\multicolumn{3}{c}{\textbf{MLLM-based fine-tuned metrics (Top-3)}} \\
\midrule
\faUnlock\; VideoPhy-2 \cite{videophy2} (Physical Commonsense)     & 0.28 & 0.37 \\
\faUnlock\; VideoPhy-2 \cite{videophy2} (Semantic Adherence)       & 0.19 & 0.16 \\
\faUnlock\; VideoScore2 \cite{videoscore2} (Physical Consistency) & 0.18 & 0.17 \\
\midrule
\multicolumn{3}{c}{\textbf{MLLM Prompting (Top-3)}} \\
\midrule
\faLock\; GPT-5 \cite{gpt5}                       & \underline{0.45} & \underline{0.38} \\
\faLock\; Gemini-2.5-Pro \cite{gemini2p5}            & 0.31 & 0.26 \\
\faLock\; GPT-4o \cite{gpt4o}                      & 0.34 & 0.31 \\
\midrule
\multicolumn{3}{c}{\textbf{Ours}} \\
\midrule
{\actionname} {\actionscore} (Ours)    & \textbf{0.61} & 0.45 \\
{\tempname} {\tempscore} (Ours)          & 0.53 & \textbf{0.64} \\
\midrule
$\Delta$ over best baseline & + 0.16 & + 0.26 \\
Relative improvement (\%) over best baseline & + 35.6\% & + 68.4\% \\
\midrule
\multicolumn{3}{c}{\emph{Inter-rater agreement}} \\
\midrule
Human vs Human & 0.72 & 0.71 \\
\bottomrule
\end{tabular}
}
\caption{\textbf{Correlation (Spearman's $\rho$) between model predictions and human scores for {\actionname} and {\tempname}.} (Higher is better). `VideoMAE(UCF101)-classification' uses the confidence score ~\cite{tong2022videomae} as the predicted scores. \faUnlock\; denotes open-source models, while \faLock\; denotes closed-source models. We observe that the proposed {\actionscore} outperforms all methods for {\actionname}, and  {\tempscore} for {\tempname}. The next best performing metric is \underline{underlined}. We report the top-3 performing baselines for each category; more results in suppl.}
\label{tab:baselines}
\vspace{-2.5em}
\end{table}

We outline the model implementation and training setup in Sec.~\ref{sec:implementation_details}. Next, we evaluate our proposed metrics against existing video evaluation metrics on TAG-Bench-v0 (Sec~\ref{sec:baselines}) and external benchmarks (Sec~\ref{sec:vbench}). We then compare multiple open- and closed-source video generative models in Sec.~\ref{sec:compare}, followed by an analysis on key design choices underpinning our approach in Sec.~\ref{sec:abl}.

\subsection{Implementation details} 
\label{sec:implementation_details}

\noindent\textbf{Features:} We extract SMPL parameters using TokenHMR~\cite{dwivedi2024tokenhmr} (Sec.~\ref{sec:3d_feats}) and 2D keypoints using DWPose~\cite{dwpose} (Sec.~\ref{sec:2d_feats}). TokenHMR employs Detectron2~\cite{wu2019detectron2} to obtain bounding boxes for the person; the cropped image is then used to infer SMPL parameters (Sec.~\ref{sec:3d_feats}). Visual appearance features (Sec.~\ref{sec:app_feats}) are obtained from the frozen ViT-H/16 backbone of TokenHMR. Motion features (Sec.~\ref{sec:motion_feats}) are computed as frame-to-frame differences in feature values. Specifically, for pose and global orientation (rotation matrices), we compute relative rotations between adjacent frames to capture angular motion, while for appearance, body shape, and 2D keypoints, we use Euclidean ($\ell_2$) distance. All features are flattened and normalized prior to being passed to the model.

\noindent\textbf{Model details:} Recall that each feature is first processed by a $1$D temporal convolutional block (Sec.~\ref{sec:model_training}), which contains three sequential $1$D convolution layers with kernel size $5$ and respective dilation factors $\{1, 2, 4\}$. The dilated convolutions enable the model to efficiently capture both short- and mid-range temporal patterns. A residual connection is applied after each layer~\cite{resnet}. Each temporal convolutional block outputs a $256$-dimensional embedding per input, which are then fused using an attention weighing mechanism to produce a single $256$-D representation per frame.
Next, a learnable $256$-D $\text{[CLS]}$ token is prepended to the sequence of $256$-D frame representations, and sinusoidal positional embeddings are added. This sequence is processed by a $4$-layer Transformer encoder with $8$ attention heads, resulting in $\mathbf{z}_{\text{CLS}}$ and $\{\mathbf{z}_t\}_{t=1}^{T}$ which are then $\ell_2$-normalized.

\noindent\textbf{Training data:} We train our encoder from scratch using the same 10 UCF101 action categories mentioned in the human evaluation (Sec.~\ref{sec:human_eval}). We use videos at their native resolution ($320 \times 240$) and frame rate ($25$ FPS), and divide into temporal windows of $T{=}32$ frames with an overlap of $8$ frames between two windows. We exclude videos whose first frame was used as input for generating videos for TAG-Bench-v0 (Sec.~\ref{sec:human_eval}). To extract only features of the human performing the action, we further discard videos containing more than one person. The remaining real videos are split into training ($80\%$) and validation ($20\%$).

\noindent\textbf{Training details.} We train the encoder for $90$ epochs using AdamW~\cite{adamw} with a learning rate of $3{\times}10^{-4}$, weight decay of $1{\times}10^{-4}$, batch size $256$, $\lambda{=}10$ in Eq.~\ref{eq:loss_total}, and a cosine learning rate schedule, on a single A100 GPU. 

\noindent\textbf{Evaluation:} Following prior work~\cite{videoscore, videoscore2}, we report Spearman’s rank correlation ($\rho$) between the metrics and human ratings.


\subsection{Comparison to automatic metrics and MLLMs}
In this section, we evaluate a diverse set of automatic video quality metrics on TAG-Bench-v0 (Sec.~\ref{sec:human_eval}), including feature-based methods, fine-tuned MLLM evaluators, and zero-shot approaches, assessing their alignment with human ratings of action correctness and motion quality (Table~\ref{tab:baselines}).
\label{sec:baselines}

\noindent\textbf{Feature-based metrics:}
Frame-level metrics like CLIP-sim~\cite{clip} and DINO-dim~\cite{dino} fail to capture motion dynamics, correlating poorly with human ratings ($<\!0.22$). We also note that the metrics from VBench-2.0~\cite{vbench2} designed to assess human anatomy and identity per frame also show very low correlations on our benchmark ($<0.11$) in the Appendix.


\begin{figure*}[t]
\centering

\begin{minipage}[t]{0.64\linewidth}
    \subcaptionbox{\actionname\ (TAG-Bench-v0)}[0.49\linewidth]
    {\includegraphics[width=\linewidth]{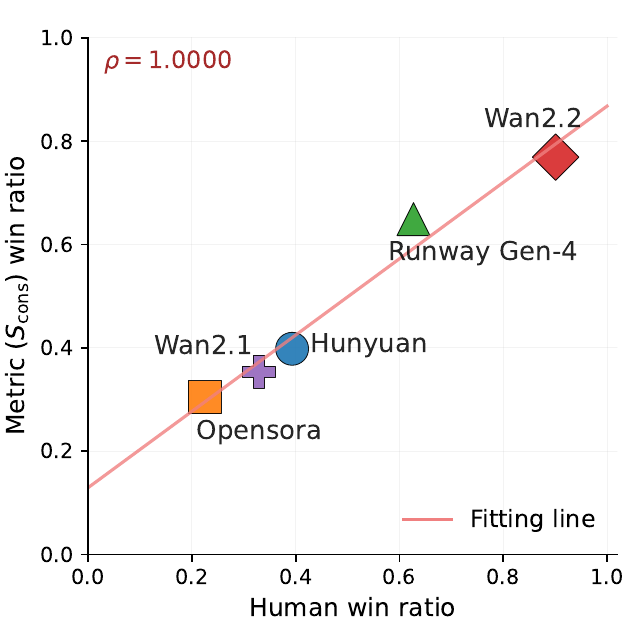}}
    \hfill
    \subcaptionbox{\tempname\ (TAG-Bench-v0)}[0.49\linewidth]
    {\includegraphics[width=\linewidth]{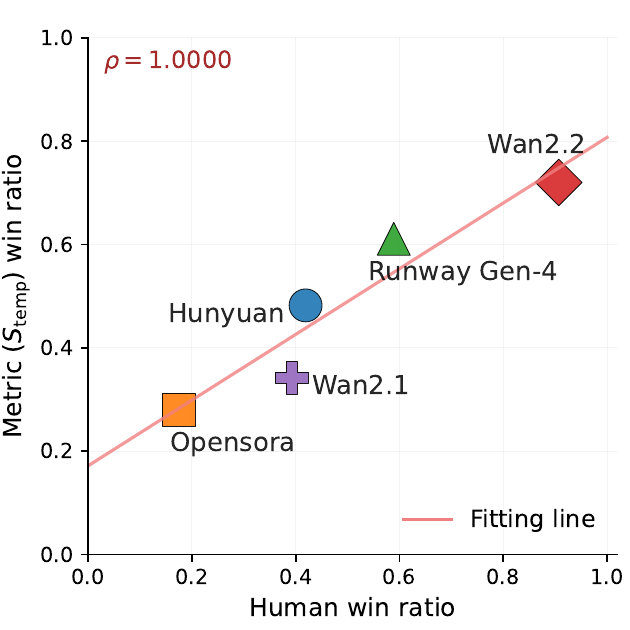}}
\end{minipage}
\hfill
\begin{minipage}[t]{0.32\linewidth}
    \subcaptionbox{\tempname\ (VBench-2.0)}
    {\includegraphics[width=\linewidth]{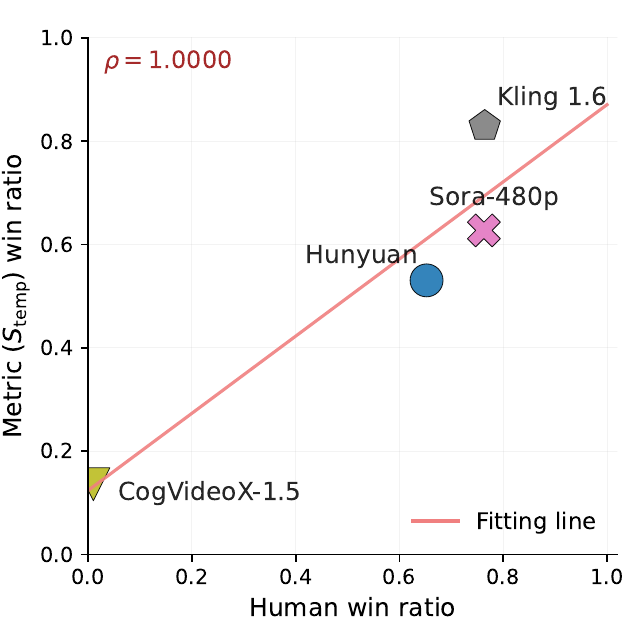}}
\end{minipage}
\vspace{-1.0em}
\caption{\textbf{Model comparisons on TAG-Bench-v0 and VBench-2.0 Human Anatomy.}
We compare models pairwise for the same input prompt; for each pair, the model with the higher score (human or metric) is the winner.
We then plot the win ratios (see \cref{sec:vbench}) of human scores (x-axis) against win ratios from our metric (y-axis).
Our metrics ($\actionscore$ and $\tempscore$) observe the same ranking of models as humans on both benchmarks.}
\label{fig:winrate}
\vspace{-1.0em}
\end{figure*}

\noindent\textbf{MLLM-based fine-tuned evaluators:}
Recent methods like VideoScore2~\cite{videoscore, videoscore2} and VideoPhy-2~\cite{videophy2} fine-tune MLLMs on human ratings to predict video quality. However, they target general criteria such as visual fidelity or text–video alignment, rather than fine-grained human motion. For instance, the \emph{Physical Commonsense} score in VideoPhy-2 assesses scene- and object-level physics (e.g., gravity, collisions), but not human-specific dynamics such as joint coordination. As a result, its alignment with human ratings on our benchmark is modest, achieving only $0.28$ on {\actionname} and $0.37$ on {\tempname}.

\noindent\textbf{Prompting MLLMs:}
We assess how well existing MLLMs align with human judgments. 
Directly using raw video inputs proved unreliable, as MLLMs often fail to capture fine-grained temporal details in human motion~\cite{xue2025seeing}.
For instance, Gemini-2.5-Pro~\cite{gemini2p5}
shows low correlations of $0.25$ for {\actionname} and $0.22$ for {\tempname}.
To mitigate this and provide more direct visual evidence, we uniformly sample $40$ frames from each video and arrange them into $4{\times}10$ grid panels (suppl.).
This layout preserves both temporal progression and spatial structure, offering clearer visual evidence.
Under this setting, we find a 
stronger alignment with human judgments: e.g., Gemini-2.5-Pro’s correlations with human scores is $0.31$ on \textit{{\actionname}} and  $0.26$ on \textit{{\tempname}}. Among MLLMs, GPT-5 achieves the highest alignment, with correlations of $0.45$ for {\actionname} and $0.38$ for {\tempname} (Table~\ref{tab:baselines}).

\noindent\textbf{Proposed metrics:} Our metrics (Sec.~\ref{sec:embedding_metrics}) show stronger alignment with human perception. {\actionname} ({\actionscore}) achieves a correlation of $0.61$ while {\tempname} ({\tempscore}) achieves $0.64$. 
Notably, despite being trained on a much smaller dataset and using only a few features, our metrics outperform GPT-5 by $\mathbf{+35.6\%}$ and $\mathbf{+68.4\%}$ in relative gains, respectively. Both correlations are statistically significant ($p < 0.05$). The 95\% bootstrap confidence intervals over the $300$ TAG-Bench-v0 videos are $[0.52, 0.63]$ and $[0.59, 0.68]$ for {\actionscore} and {\tempscore}, respectively, indicating the statistical stability of our metrics.

\subsection{Performance on external benchmarks}
\label{sec:vbench}

We evaluate on the Human Anatomy subset of VBench-2.0~\cite{vbench2}, which compares videos from four \emph{text-to-video} models (Sora-480p~\cite{openai2024sora}, Kling~\cite{kling2024}, Hunyuan~\cite{kong2024hunyuanvideo}, and CogVideo~\cite{hong2022cogvideo}) on a fixed set of text prompts designed to expose anomalies in human appearance and structure in generated videos (e.g, ``A woman is cutting objects"). Given these four models, annotators select the more realistic video in pairwise comparisons for each prompt.
Following Sec.~\ref{sec:implementation_details}, we evaluate only videos with a single visible person per frame, and compute \textit{win ratios}~\cite{vbench2}. A model ``wins” when its video is preferred by annotators, and its win ratio is the fraction of total comparisons it wins (i.e. number of wins / total pairwise comparisons). Models are ranked by these ratios (higher is better).
We then compare these human-derived rankings to those inferred by
our {\tempscore} metric, by computing win ratio in a similar fashion.
Since VBench-2.0 prompts do not correspond to the $10$ classes used in training (Sec.~\ref{sec:implementation_details}), we evaluate using only {\tempscore}, as {\actionscore} requires a corresponding action centroid. As shown in Fig.~\ref{fig:winrate}, {\tempscore} produces the same model ranking as human raters. 
Notably, the evaluated videos from VBench-2.0 are generated by \emph{text-to-video} models and are prompted with actions not present in our training set (Sec.~\ref{sec:implementation_details}).
This demonstrates that the learned motion-sensitive embedding generalizes beyond the plausibility learned from actions seen during training.

\subsection{Comparing generative models}
\label{sec:compare}
Having shown that our metrics correlate strongly with human judgments (Sec.\ref{sec:baselines}), we now use them for a fine-grained comparison of the five state-of-the-art generative models evaluated in our study (Sec.\ref{sec:human_eval}).


\begin{figure}[!t]
\centering

\begin{minipage}{0.49\linewidth}
  \centering
  \includegraphics[width=\linewidth]{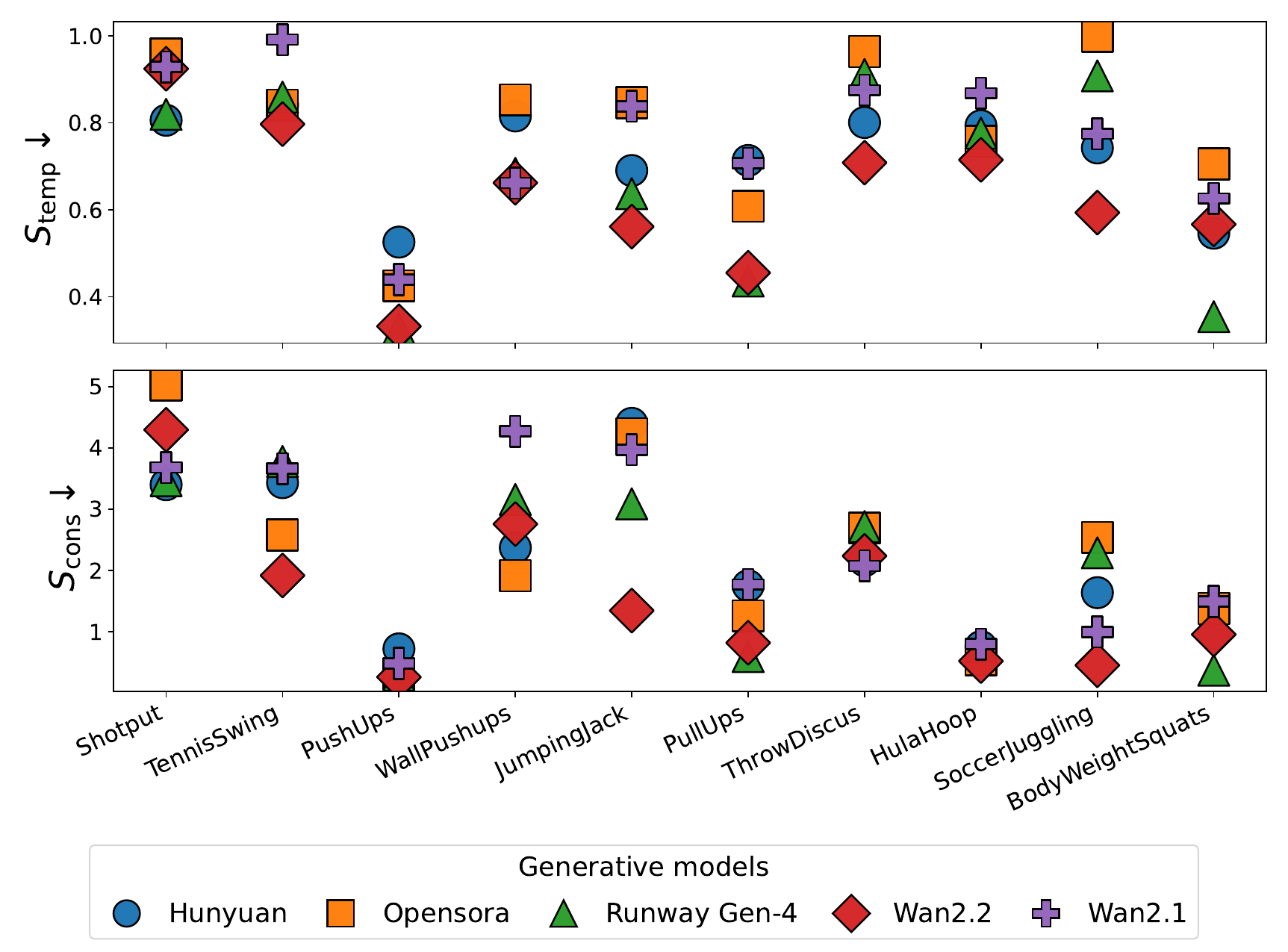}
  \vspace{-2.0em}
  \captionof{figure}{\textbf{Comparing generative models.} We plot the mean {\actionscore} and {\tempscore} scores (Sec.~\ref{sec:embedding_metrics}) (lower is better) for each generative model across different actions. Wan2.2 performs best among the other models (low scores in both {\actionscore} and {\tempscore}). \textit{Shotput} and \textit{JumpingJack} challenge all models, yielding high scores across both metrics.}
  \label{fig:compare}
\end{minipage}
\hfill
\begin{minipage}{0.49\linewidth}
  \centering
  \includegraphics[width=\linewidth]{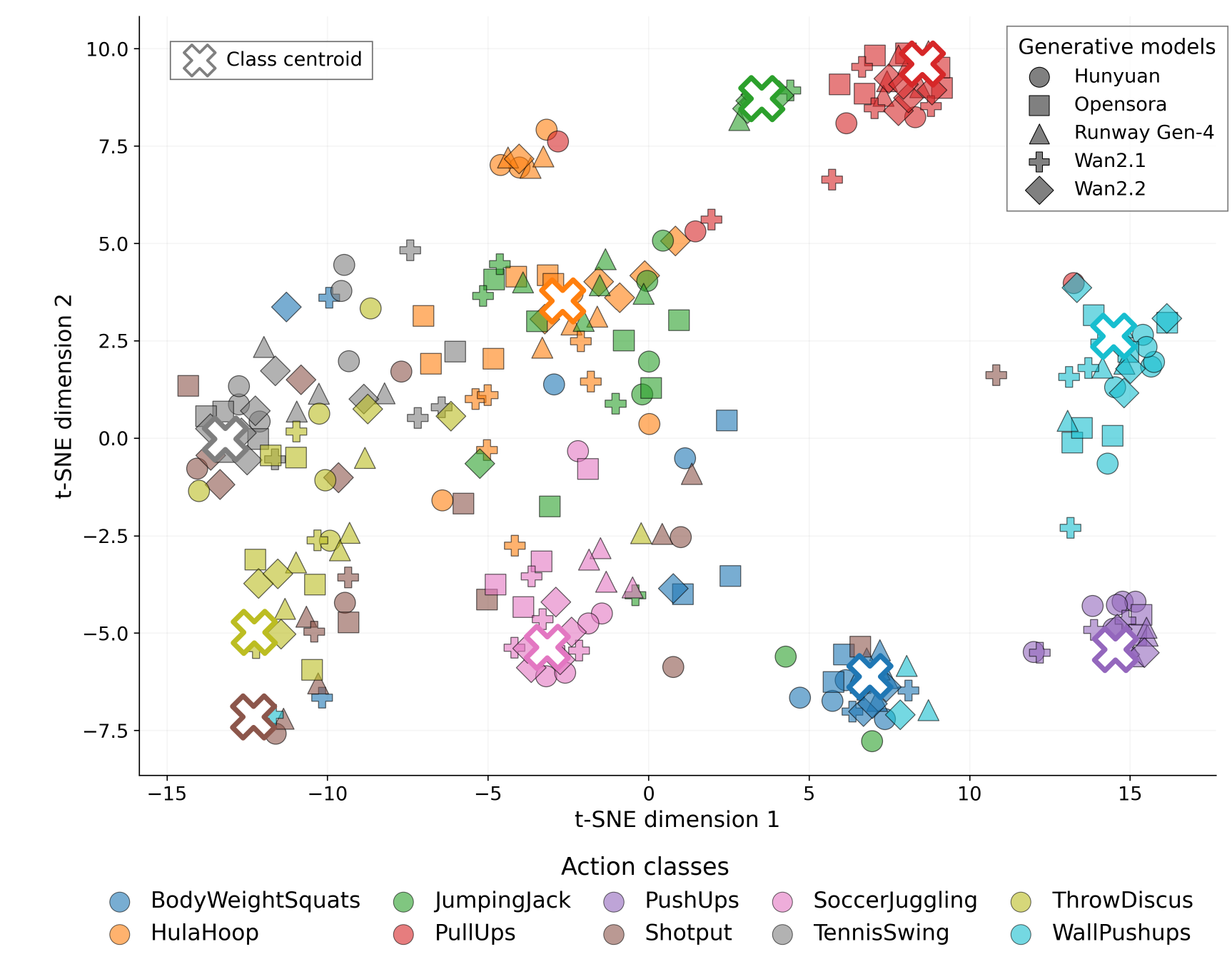}
  \vspace{-2.0em}
  \captionof{figure}{\textbf{t-SNE visualization of the embeddings of generated videos along with train centroids.} We project the $z_{\text{CLS}}$ embeddings of \emph{generated} videos (colored markers) from TAG-Bench-v0 and the corresponding \emph{training} class centroids (white crosses) using t-SNE~\cite{tsne}. Realistic generated videos cluster near their respective class centroids (e.g., Wan2.2 videos for ``PullUps", with an average human rating of: $\mathbf{8.41}$ for {\actionname}), while poorly generated videos lie further away (e.g., Wan2.2 videos for ``Shotput" with an average human rating of: $\mathbf{4.43}$) (Sec.~\ref{sec:human_eval}).}
  \label{fig:tsne_clusters}
\end{minipage}
\vspace{-0.5em}
\end{figure}
\noindent\textbf{Overall model performance (win ratios).} 
We compare all models on TAG-Bench-v0 using win ratios (Sec.~\ref{sec:vbench}).
As shown in Fig.~\ref{fig:winrate}, Wan2.2~\cite{wan2025wan} achieves the highest win ratios ($0.77$ for {\actionname}, $0.72$ for {\tempname}), outperforming the closed-source Runway Gen-4~\cite{runway2024gen4} ($0.65$ and $0.61$, respectively). 

\noindent \textbf{Which actions are difficult to generate for which model?}
As shown in Fig.~\ref{fig:compare}, Wan2.2 consistently achieves the lowest (best) {\actionscore} and {\tempscore} for most actions. 
However, a video may depict the correct action while showing unrealistic motion.
For instance, Wan2.2 performs well on ``SoccerJuggling'' in {\actionscore} but relatively poorly in {\tempscore}.
This gap aligns with human ratings\footnote{We report the average metric scores on the original 1–10 scale, computed across valid human raters (Sec.~\ref{sec:human_eval}) for each video.}, where the human scores are $8.4$ for {\actionname} and $7.6$ for {\tempname}, suggesting that viewers perceived the action as semantically correct but noticeably less natural. In contrast, Runway-Gen4 surpasses all models on ``BodyWeightSquats'' for both metrics. \underline{Thus, no single model dominates across all actions.}
\\

\noindent \textbf{Are some actions universally harder to generate?} From Fig.~\ref{fig:compare}, we note that 
actions such as ``Shotput'' and ``ThrowDiscus'' are challenging across all five models, which show high (i.e., poor) {\actionscore} and {\tempscore} scores in Fig.~\ref{fig:compare}. This suggests that complex, full-body rotational actions remain a common failure case for current generative models. In contrast, actions involving less dynamic motion such as ``PushUps'' and ``PullUps'' are relatively easier to generate, as reflected by low scores in both metrics. This trend is illustrated in the t-SNE~\cite{tsne} visualization in Fig.~\ref{fig:tsne_clusters}: videos with higher human and automatic ratings (i.e. higher quality) cluster tightly around their class centroids (e.g., ``PullUp''), while lower-quality generations appear dispersed away (e.g., ``Shotput'').



\subsection{Analysis}

\label{sec:abl}

\begin{table}[!t]
\centering
\setlength{\tabcolsep}{3pt}
\renewcommand{\arraystretch}{0.95}

\begin{minipage}{0.48\linewidth}
\centering
\resizebox{0.68\linewidth}{!}{%
\begin{tabular}{cccc}
\toprule
$\mathcal{L}_{\text{supcon}}$ & $\mathcal{L}_{\text{hard-neg}}$ &
\makecell{\textbf{Action}\\\textbf{Consistency}} &
\makecell{\textbf{Temporal}\\\textbf{Coherence}} \\
\midrule
\cmark & \cmark & \textbf{0.61} & \textbf{0.64} \\
\xmark & \cmark & 0.26 & 0.38 \\
\cmark & \xmark & 0.54 & 0.57 \\
\bottomrule
\end{tabular}%
}
\caption{\textbf{Effect of the loss terms} $\mathcal{L}_{\text{supcon}}$ and $\mathcal{L}_{\text{hard-neg}}$. Both terms jointly improve action consistency and temporal coherence; removing either hurts.}
\label{tab:loss_terms}
\end{minipage}
\hfill
\begin{minipage}{0.48\linewidth}
\centering
\resizebox{0.98\linewidth}{!}{%
\begin{tabular}{cccccccc}
\toprule
\textbf{Pose} &
\makecell{\textbf{Body}\\\textbf{shape}} &
\makecell{\textbf{Global}\\\textbf{orientation}} &
\textbf{Keypoints} &
\makecell{\textbf{Visual}\\\textbf{features}} &
\textbf{Motion} &
\makecell{\textbf{Action}\\\textbf{Consistency}} &
\makecell{\textbf{Temporal}\\\textbf{Coherence}} \\
\midrule
\cmark & \cmark & \cmark & \cmark & \cmark & \cmark & \textbf{0.61} & \textbf{0.64} \\
\xmark & \cmark & \cmark & \cmark & \cmark & \cmark & 0.56 & 0.57 \\
\cmark & \xmark & \cmark & \cmark & \cmark & \cmark & 0.54 & 0.57 \\
\cmark & \cmark & \xmark & \cmark & \cmark & \cmark & 0.57 & 0.57 \\
\cmark & \cmark & \cmark & \xmark & \cmark & \cmark & 0.61 & 0.57 \\
\cmark & \cmark & \cmark & \cmark & \xmark & \cmark & 0.56 & 0.59 \\
\cmark & \cmark & \cmark & \cmark & \cmark & \xmark & 0.46 & 0.50 \\
\bottomrule
\end{tabular}%
}
\caption{
\textbf{Effect of each input feature.} We report Spearman’s correlation ($\rho$) with human scores after zeroing each input feature independently. Models are retrained from scratch for each setting. “Motion’’ denotes temporal derivatives of all inputs (\cref{sec:motion_feats}). Removing motion causes the largest degradation.}
\label{tab:input_ablation}
\end{minipage}

\vspace{-2.5em}
\end{table}
\noindent\textbf{Effect of loss terms.}
We evaluate the contribution of each loss term in Eq.\ref{eq:loss_total} by training variants of the encoder with specific losses removed.
From Table~\ref{tab:loss_terms}, we observe that removing the supervised contrastive loss ($\mathcal{L}_{\text{supcon}}$) causes a steep drop in {\actionname} (from $0.61~{\rightarrow}~\textcolor{Red}{0.26}$), validating that $\mathcal{L}_{\text{supcon}}$ is essential for structuring the embedding space according to action semantics. The addition of $\mathcal{L}_{\text{hard-negative}}$ further refines this embedding space, improving both {\actionname} ($0.54~{\rightarrow}~\textcolor{ForestGreen}{0.61}$) and {\tempname} ($0.57~{\rightarrow}~\textcolor{ForestGreen}{0.64}$). $\mathcal{L}_{\text{supcon}}$ also proves crucial for {\tempname} (improving scores from $0.38~{\rightarrow}~\textcolor{ForestGreen}{0.64}$). These results suggest that learning the rules of plausible motion (\emph{how}) is most effective when the embedding space encodes meaningful action semantics (\emph{what}). 

\noindent\textbf{Ablation study on input features.} To understand the importance of each input feature, we retrain the encoder while zeroing out one input feature at a time (e.g., setting the pose features to zero, while retaining the rest). This retains the full dimensionality of the input and does not alter model capacity. As shown in Table~\ref{tab:input_ablation}, masking any single feature leads to a decrease in performance. For instance, masking $3D$ pose results in a drop in correlation scores for {\actionname} from $0.61~{\rightarrow}$~\textcolor{Red}{$0.56$}. Masking all motion features results in the largest drop ({\actionname}: $0.61~{\rightarrow}$~\textcolor{Red}{$0.46$}). This highlights the necessity of our multi-feature static and motion representations. 



\newcommand{\clipdinoScale}{0.6}  

\begin{table}[t]
\centering
\small
\renewcommand{\arraystretch}{1.00}


\begin{subtable}[b]{0.49\linewidth} 
\centering
\scalebox{\clipdinoScale}{%
\begin{tabular}{@{}lcc@{}}
\toprule
\shortstack{\textbf{Visual}\\\textbf{features}} &
\shortstack{\textbf{Action}\\\textbf{Consistency}} &
\shortstack{\textbf{Temporal}\\\textbf{Coherence}} \\
\midrule
ViT (TokenHMR)  & \textbf{0.61} & \textbf{0.64} \\
CLIP            & 0.51 & 0.39 \\
DINOv2          & 0.56 & 0.38 \\
\bottomrule
\end{tabular}}
\caption{With human-centric features}
\end{subtable}
\hfill
\begin{subtable}[b]{0.49\linewidth} 
\centering
\scalebox{\clipdinoScale}{%
\begin{tabular}{@{}lcc@{}}
\toprule
\shortstack{\textbf{Visual}\\\textbf{features}} &
\shortstack{\textbf{Action}\\\textbf{Consistency}} &
\shortstack{\textbf{Temporal}\\\textbf{Coherence}} \\
\midrule
CLIP-only features    & 0.12 & 0.19 \\
DINOv2-only features  & 0.14 & 0.26 \\
\bottomrule
\end{tabular}}
\caption{Without human-centric features}
\end{subtable}

\vspace{-1.0em}
\caption{\textbf{Which visual feature backbone helps the most?} (a) Replacing ViT features (from TokenHMR~\cite{dwivedi2024tokenhmr}) with CLIP~\cite{clip} or DINOv2~\cite{Oquab2023DINOv2LR} embeddings of the person cropped from each frame reduces correlation with human judgments. (b) Using CLIP or DINOv2 embeddings alone (without the human-centric 3D or 2D features described \cref{sec:human_features}) as inputs to the encoder performs worst, validating that structured human-centric features are essential for our task.}
\label{tab:clip_dino}
\vspace{-2.0em}
\end{table}

\noindent\textbf{Effect of different visual appearance features.} To assess different choices for the visual appearance feature $f_{\text{vis}}$, we train the encoder from scratch by replacing the ViT features from TokenHMR~\cite{dwivedi2024tokenhmr} with CLIP~\cite{clip} and DINOv2~\cite{Oquab2023DINOv2LR} embeddings extracted from the cropped person images (Sec.~\ref{sec:implementation_details}), while keeping all other components fixed. As shown in Table~\ref{tab:clip_dino}, replacing ViT features with CLIP or DINOv2 lowers performance (e.g., {\actionname}: $0.61~{\rightarrow}$\textcolor{Red}{~$0.51$}, {\tempname}: $0.64~{\rightarrow}$\textcolor{Red}{~$0.39$}, when replaced with CLIP features). 
Furthermore, training the encoder using only CLIP or DINOv2 features (i.e., without any human-centric 3D or 2D features) further degrades performance, confirming that our strong results stems not only from the model design and data, but also from the integration of specialized human-centric representations.

\noindent\textbf{Extending TAG-Bench to more action classes.} To assess the generalization of our framework, we extend TAG-Bench-v0 from $10$ to $23$ action classes by incorporating $13$ additional actions: \textit{Bowling, Clean\&Jerk, GolfSwing, HammerThrow, Hammering, HandStandPushup, JugglingBalls, JumpRope, Lunges, PlayingGuitar, RockClimbingIndoor, RopeClimbing,} and \textit{Surfing}. We follow the same human evaluation protocol as in Sec.~\ref{sec:human_eval} for the additional classes. Our model when trained and evaluated on all $23$ classes continues to outperform -- correlation scores of {\actionname}: $0.59$ and {\tempname}: $0.56$, compared to GPT-5's scores of {\actionname}: $0.46$ and {\tempname}: $0.39$.


\section{Discussion and Future Work}
We present a framework for evaluating human actions in generated videos by decomposing the task into two aspects: action consistency against real videos and the temporal smoothness of human anatomy. We identify a critical gap affecting \textit{all} current benchmarks and metrics for this evaluation. To address this, we introduce \textbf{TAG-Bench-v0} and two new metrics, which demonstrate strong alignment with human perceptual judgment and generalizability across diverse generation models. While our main study is conducted on $10$ action classes, it offers a proof-of-concept for the utility of the proposed features.  Sec.~\ref{sec:abl} shows the extensibility of the framework to more actions. Future work will extend this framework to more actions, longer-form videos, and in-depth exploration of integrating human-physics-based features with modern MLLMs.
\section*{Acknowledgements}
This research was supported by Meta. We thank Sai Kumar Dwivedi, Jonathan Granskog, Ed Chien, Robin Kahlow, Lorenzo Torresani, Xingjian (Jessie) Han, and members of our research group at BU for helpful discussions and feedback. We also thank Runway for providing us computational resources for this project.

\newpage

%
%

\bibliographystyle{splncs04}
\bibliography{main}

\clearpage
\setcounter{page}{1}
\appendix
\phantomsection

\begin{center}
  {\large \textbf{Generative Action Tell-Tales: Assessing Human Motion in Synthesized Videos}\\}
  \vspace{1ex}
  {\large Supplementary Material\\}
\end{center}

\phantomsection
\section*{Table of Contents}

\begin{tabular}{@{}p{0.06\linewidth}p{0.88\linewidth}@{}}
\textbf{A.} & \hyperref[sec:human_eval_ui]{Human Evaluation User Interface (UI)} \dotfill \pageref{sec:human_eval_ui} \\[0.3ex]

\textbf{B.} & \hyperref[sec:tagbench]{TAG-Bench-v0 curation} \dotfill \pageref{sec:tagbench} \\
& \quad \hyperref[subsec:model_configs]{B.1 Model configurations} \dotfill \pageref{subsec:model_configs} \\
& \quad \hyperref[subsec:subject_rejection]{B.2 Subject rejection} \dotfill \pageref{subsec:subject_rejection} \\
& \quad \hyperref[subsec:convergence]{B.3 Convergence analysis of human ratings} \dotfill \pageref{subsec:convergence} \\[0.3ex]

\textbf{C.} & \hyperref[sec:impl_details]{Additional implementation details} \dotfill \pageref{sec:impl_details} \\
& \quad \hyperref[subsec:tokenhmr]{C.1 TokenHMR feature extraction} \dotfill \pageref{subsec:tokenhmr} \\
& \quad \hyperref[subsec:train_processing]{C.2 Training data processing} \dotfill \pageref{subsec:train_processing} \\
& \quad \hyperref[subsec:human_feats]{C.3 Human-centric features extracted from each frame} \dotfill \pageref{subsec:human_feats} \\[0.3ex]

\textbf{D.} & \hyperref[sec:win_ratios]{Win Ratios (TAG-Bench-v0, VBench-2.0)} \dotfill \pageref{sec:win_ratios} \\[0.3ex]

\textbf{E.} & \hyperref[sec:addl_experiments]{Additional Experiments} \dotfill \pageref{sec:addl_experiments} \\
& \quad \hyperref[subsec:act_class_sep]{E.1 Action class separation in the embedding space} \dotfill \pageref{subsec:act_class_sep} \\
& \quad \hyperref[subsec:temp_dist]{E.2 Sensitivity to temporal distortions} \dotfill \pageref{subsec:temp_dist} \\
& \quad \hyperref[subsec:window_exp]{E.3 Temporal window length ($T$)} \dotfill \pageref{subsec:window_exp} \\
& \quad \hyperref[subsec:outlier_exp]{E.4 Visualizing outliers from Fig.~\ref{fig:tsne_clusters}} \dotfill \pageref{subsec:outlier_exp} \\
& \quad \hyperref[subsec:llm_exp]{E.5 LLM-enhanced prompts for video generation} \dotfill \pageref{subsec:llm_exp} \\
& \quad \hyperref[subsec:gen_neg_exp]{E.6 Generated videos as hard-negatives} \dotfill \pageref{subsec:gen_neg_exp} \\
& \quad \hyperref[subsec:train_exp]{E.7 Training on more data} \dotfill \pageref{subsec:train_exp} \\
& \quad \hyperref[subsec:attentn_weights]{E.8 Attention weights} \dotfill \pageref{subsec:attentn_weights} \\
& \quad \hyperref[subsec:arch_design]{E.9 Design choices} \dotfill \pageref{subsec:arch_design} \\
& \quad \hyperref[subsec:num_samples]{E.10 Sensitivity to number of samples} \dotfill \pageref{subsec:num_samples} \\[0.3ex]

\textbf{F.} & \hyperref[sec:baseline_F]{Baselines} \dotfill \pageref{sec:baseline_F} \\
& \quad \hyperref[subsec:baseline_f1]{F.1 Feature-based metrics} \dotfill \pageref{subsec:baseline_f1} \\
& \quad \hyperref[subsec:baseline_f2]{F.2 MLLM based metrics} \dotfill \pageref{subsec:baseline_f2} \\[0.3ex]

\textbf{G.} & \hyperref[sec:mllm_eval]{MLLM Prompting} \dotfill \pageref{sec:mllm_eval} \\
& \quad \hyperref[subsec:mllm_prompt]{G.1 Prompt used for MLLM evaluation} \dotfill \pageref{subsec:mllm_prompt} \\
& \quad \hyperref[subsec:mllm_icl]{G.2 Limited impact of in-context learning} \dotfill \pageref{subsec:mllm_icl} \\
\end{tabular}

\newpage

\section{Human Evaluation User Interface (UI)}
\label{sec:human_eval_ui}

\begin{figure*}[h!]
    \centering
    \includegraphics[width=\linewidth]{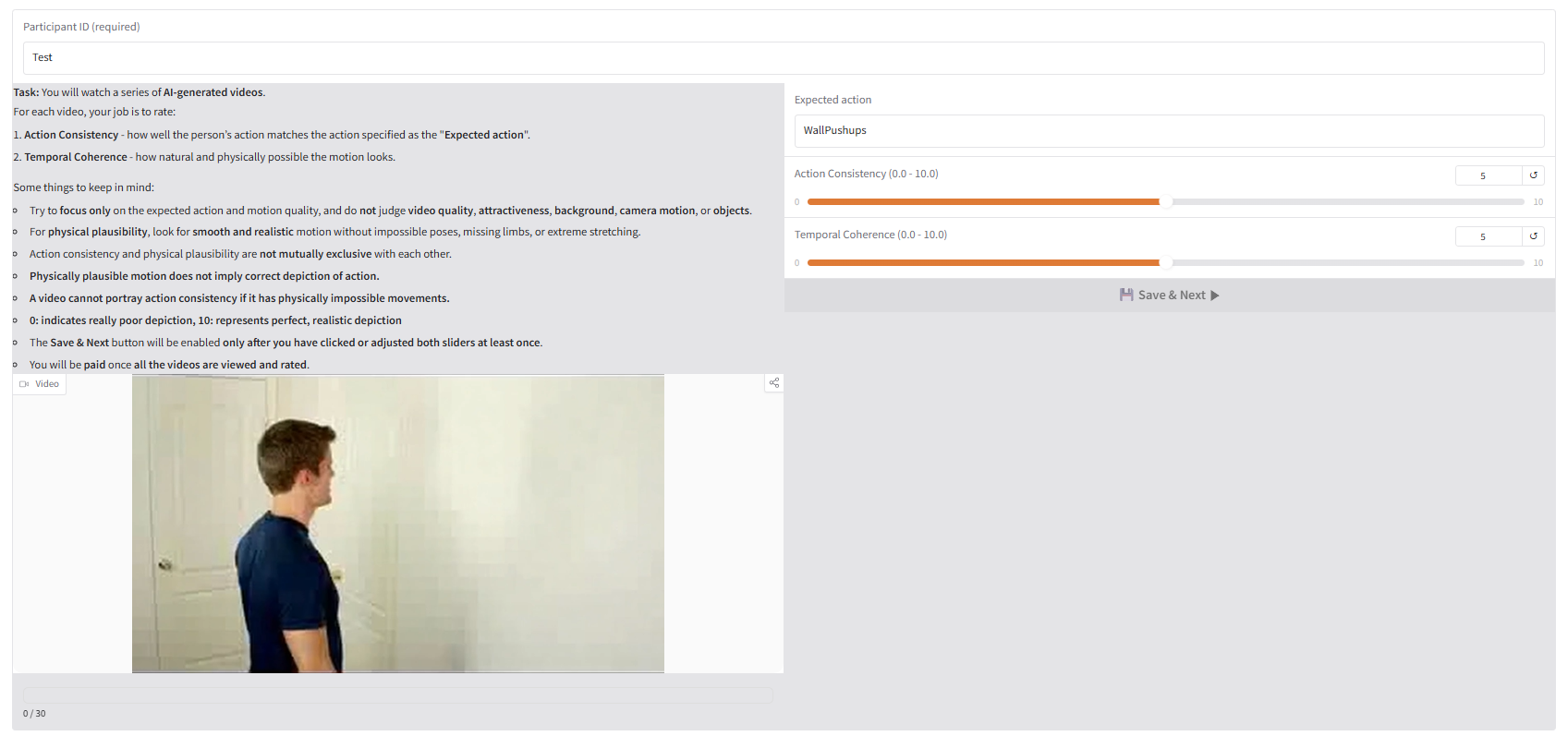}
    \caption{\textbf{User interface used for the human evaluation study.} Participants were asked to rate each AI-generated video along two dimensions: \textit{Action Consistency} (how accurately the motion matches the described action) and \textit{Temporal Coherence} (how natural and physically realistic the motion appears). Each participant viewed 30 videos and provided ratings on a scale from 0 to 10.}
    \label{fig:human_eval_ui}
\end{figure*}

\section{TAG-Bench-v0: Model configurations, Subject rejection, Aggregation of Opinion Scores}
\label{sec:tagbench}
\subsection{Model configurations}
\label{subsec:model_configs}

Below are the configurations of the image-to-video (I2V) models used to generate the videos evaluated in TAG-Bench-v0.
\begin{table}[H]
\centering
\resizebox{0.45\textwidth}{!}{%
\begin{tabular}{lcc}
\toprule
\textbf{Model} & \textbf{Resolution} & \textbf{Model name} \\
\midrule
Wan2.1~\cite{wan2025wan}      & $1104{\times}816$ & Wan2.1-I2V-14B-720P\tablefootnote{\url{https://github.com/Wan-Video/Wan2.1}} \\
Wan2.2~\cite{wan2025wan}      & $1280{\times}720$ & Wan2.2-I2V-A14B\tablefootnote{\url{https://github.com/Wan-Video/Wan2.2}} \\
Opensora~\cite{openai2024sora} & $1024{\times}576$ & Opensora-768px\tablefootnote{\url{https://github.com/hpcaitech/Open-Sora}} \\
Hunyuan~\cite{kong2024hunyuanvideo} & $1088{\times}832$ & HunyuanVideo-I2V-720p\tablefootnote{\url{https://github.com/Tencent-Hunyuan/HunyuanVideo-I2V}} \\
Runway Gen-4~\cite{runway2024gen4} & $1280{\times}720$ & Gen4-Turbo\tablefootnote{\url{https://runwayml.com/research/introducing-runway-gen-4}} \\
\bottomrule
\end{tabular}}
\caption{\footnotesize \textbf{Image-to-video (I2V) models used in TAG-Bench-v0.} Resolution indicates the native output resolution for each model variant.}
\label{tab:video_models}
\end{table}

\subsection{Subject rejection}
\label{subsec:subject_rejection}
We also select only workers with a Human Intelligence Task (HIT) approval rate of $>= 99\%$. Furthermore, to ensure that the subjective ratings are consistent and reliable, we apply a three-stage filtering process before computing the final Mean Opinion Scores (MOS) and comparing them with model predictions, described next.

\paragraph{(1) Repeated-video consistency filtering}
Among the $30$ videos shown to an annotator, $5$ are intentionally duplicated to verify response consistency; if the ratings on the duplicated videos are inconsistent, we exclude those participants. Specifically,  
for each participant, we compute the standard deviation of their ratings across the repeated samples. We retain only those participants whose scores fall within the 95th percentile on the repeated videos.
This step helps remove raters giving inconsistent scores for identical stimuli, resulting in retaining $207$ raters of the total $246$ participants.

\paragraph{(2) Subject rejection.}
Next, we apply the subject rejection procedure described in ~\cite{ITU2019BT500} to further eliminate statistically unreliable participants. The method evaluates each participant’s deviation from the population mean using two statistics:
\begin{equation}
R_1 = \frac{P_i + Q_i}{N_i}, \quad 
R_2 = \frac{|P_i - Q_i|}{P_i + Q_i},
\end{equation}
where $P_i$ and $Q_i$ are the counts of a participant’s scores that lie respectively above or below the population mean by more than the threshold (either $2S$ or $\sqrt{20}S$, depending on the distribution’s kurtosis), and $N_i$ is the number of videos rated.  Participants with $R_1 > 0.05$ and $R_2 < 0.3$ were rejected, as were those who rated fewer than ten videos.  This step resulted in rejecting $14$ additional raters, 
leaving \textbf{193 valid participants}.

\paragraph{(3) Inter-rater reliability filtering.}
Finally, we assess inter-rater reliability by computing the Spearman correlation coefficient ($\rho$) between each participant’s ratings and the aggregated mean ratings of the remaining participants.
To remove inconsistent evaluations, we excluded raters with $\rho < 0.55$;
the cutoff was set empirically to balance reliability and participant retention,
resulting in 121 retained raters for \textit{Action Consistency} and 141 for \textit{Temporal Coherence}.

After completing the three filtering stages, we computed the \textbf{MOS}
for each video along both evaluation axes—\textit{Action Consistency} and
\textit{Temporal Coherence}—as well as their \textbf{z-score normalized} versions:
\begin{equation}
z_i = \frac{x_i - \mu}{\sigma},
\end{equation}
where $x_i$ is the raw MOS of video $i$, $\mu$ is the mean MOS across all videos,
and $\sigma$ is the standard deviation. We use these as final scores throughout our work
and consistency analyses with the model evaluation results.

\subsection{Convergence analysis of human ratings}
\label{subsec:convergence}
To ensure the reliability of our human evaluations, we examined the stability of the standard deviation of scores as the number of raters increased.
For each evaluation axis (\textit{Action Consistency} and \textit{Temporal Coherence}), we selected the five videos evaluated by the largest number of participants and computed the standard deviation of their Mean Opinion Scores (MOS) in a cumulative manner, progressively adding more raters.

\begin{figure}[t]
    \centering
    \includegraphics[width=0.55\linewidth]{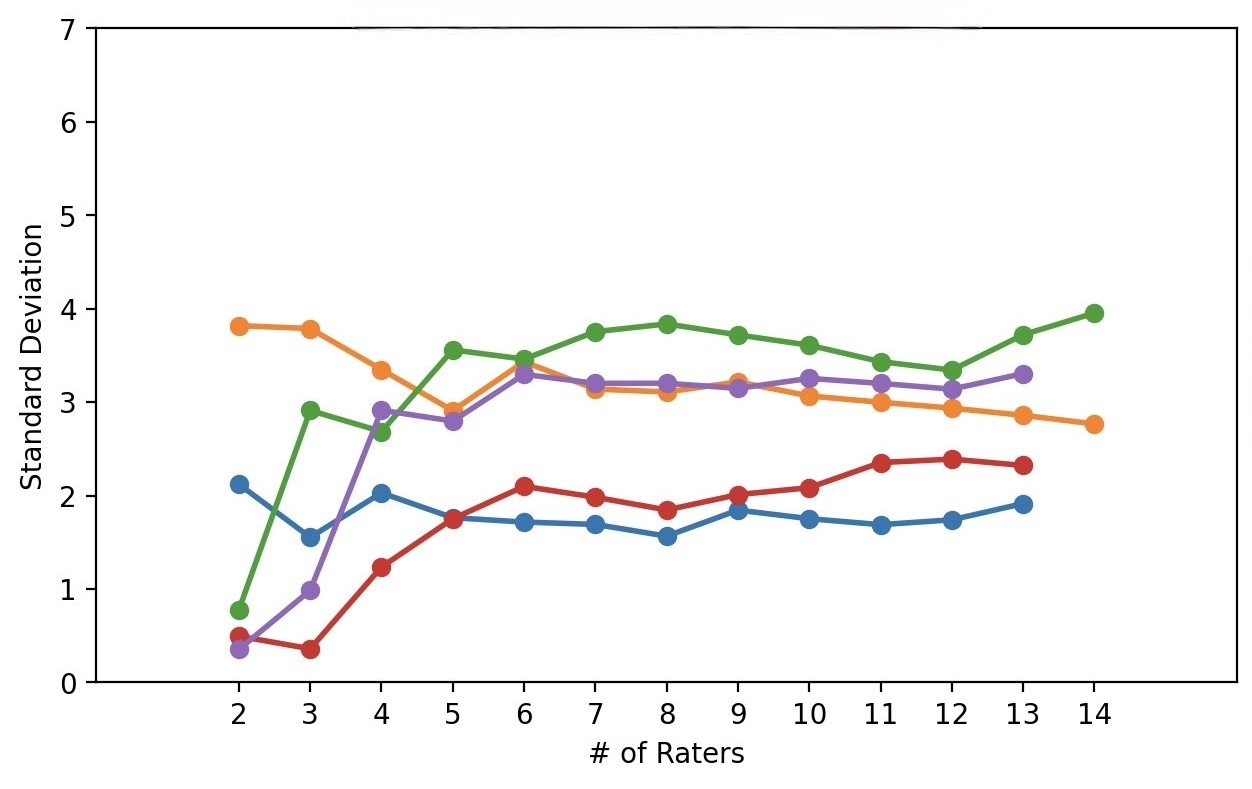}
    \vspace{-0.1in}
    \caption{Standard deviation vs. number of raters for five randomly selected videos (\textit{Action Consistency}). 
    Each curve shows the evolution of the MOS standard deviation as more raters are included.  Notice how the standard deviation is stabilizing as we add more raters.}
    \label{fig:std_ac}
\end{figure}

\begin{figure}[t]
    \centering
    \includegraphics[width=0.55\linewidth]{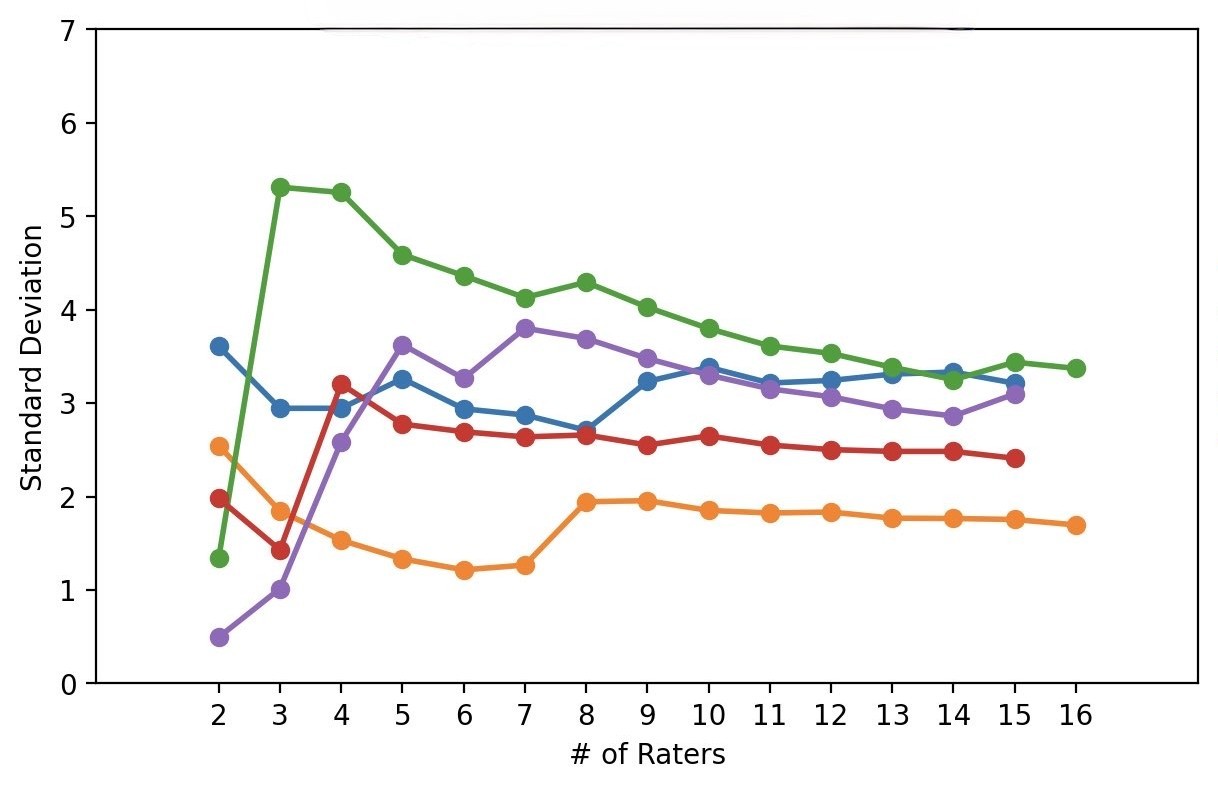}
    \vspace{-0.1in}
    \caption{Standard deviation vs. number of raters for five randomly selected videos (\textit{Temporal Coherence}). Each curve shows the evolution of the MOS standard deviation as more raters are included. Notice how the standard deviation is stabilizing as we add more raters.}
    \label{fig:std_pp}
\end{figure}

As shown in Figures~\ref{fig:std_ac} and~\ref{fig:std_pp}, the standard deviation stabilizes as the number of raters increases, typically converging after about 9--10 participants. 
This convergence indicates that the variability in human judgments remains bounded, suggesting that our collected human evaluations are statistically stable and reliable.
Hence, the aggregated MOS scores used in our main experiments are based on a sufficiently large and consistent set of raters.

\FloatBarrier
\clearpage

\section{Additional implementation details}
\label{sec:impl_details}
\subsection{TokenHMR feature extraction}
\label{subsec:tokenhmr}
\begin{figure}[h]
  \centering
  \includegraphics[width=0.45\linewidth, keepaspectratio]{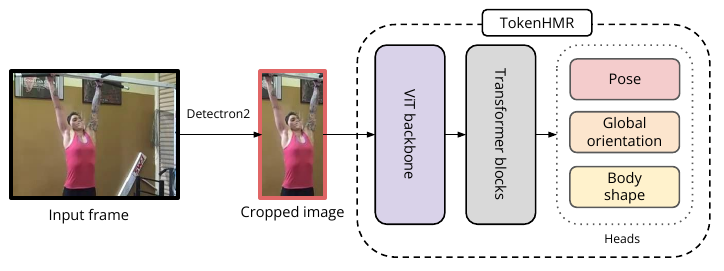}
    \caption{\footnotesize 
    \textbf{\footnotesize TokenHMR-based feature extraction.} Each input frame is processed by Detectron2~\cite{wu2019detectron2} to obtain a bounding box of the person, which is cropped and passed to TokenHMR~\cite{dwivedi2024tokenhmr}. TokenHMR uses a ViT-H/16 backbone, followed by transformer blocks and task-specific heads, to predict SMPL parameters: pose ($\theta$), global orientation ($go$), and body shape ($\beta$) in Sec.~\ref{sec:3d_feats}. Intermediate features from the ViT backbone are used as visual appearance features ($f_{\text{vis}}$) in Sec.~\ref{sec:app_feats}. We adapt the figure from TokenHMR~\cite{dwivedi2024tokenhmr}.
    }
  \label{fig:tokenhmr}
\end{figure}

\subsection{Training data processing}
\label{subsec:train_processing}

We use Detectron2~\cite{wu2019detectron2} to count the number of people in each frame and discard videos containing more than one person in any frame, as mentioned in Sec.~\ref{sec:implementation_details}. To ensure input features correspond only to the person performing action, we retain videos with only a single visible person. This filtering step yields $930$ videos out of the $1,279$ videos across the $10$ selected classes (Sec.~\ref{sec:human_eval}).

\subsection{Human-centric features extracted from each frame}
\label{subsec:human_feats}

\begin{figure}[h]
  \centering
  \includegraphics[width=0.45\linewidth, keepaspectratio]{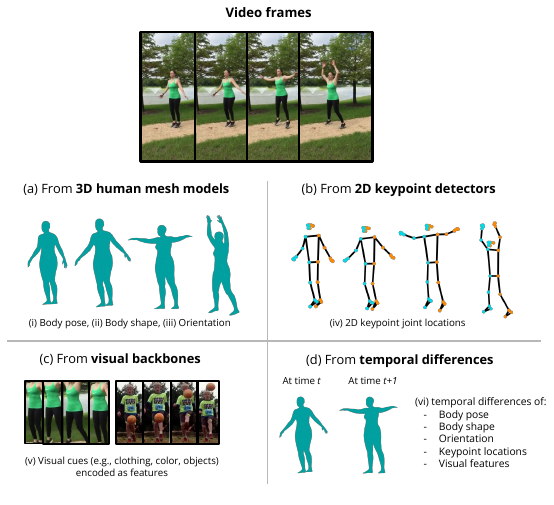}
  \caption{\footnotesize \textbf{Human-centric input features.} From each frame, we extract (a) 3D pose, body shape, and global orientation (Sec.~\ref{sec:3d_feats}), (b) 2D keypoints (Sec.~\ref{sec:2d_feats}), and (c) visual appearance features (Sec.~\ref{sec:app_feats}) to describe the body’s state in that frame. (d) We additionally compute temporal differences of each feature to capture frame-to-frame motion dynamics (Sec.~\ref{sec:motion_feats})}
  \label{fig:teal_mesh}
\end{figure}

\begin{table}[H]
\centering
\resizebox{0.55\textwidth}{!}{%
\begin{tabular}{lccccc}
\toprule
\textbf{Feature} &
\begin{tabular}{c}Pose\end{tabular} &
\begin{tabular}{c}Global\\orientation\end{tabular} &
\begin{tabular}{c}Body\\shape\end{tabular} &
\begin{tabular}{c}Keypoints\end{tabular} &
\begin{tabular}{c}Visual\\features\end{tabular} \\
\midrule
\textbf{Dimension} & $23{\times}3{\times}3$ & $1{\times}3{\times}3$ & $10$ & $60{\times}2$ & $1024$ \\
\bottomrule
\end{tabular}}
\caption{\footnotesize \textbf{Per-frame input feature dimensions.} Pose and Global orientation are $3{\times}3$ rotation matrices representing joint rotations (23 joints) and the global orientation (pelvis joint), respectively. Body shape is a $10$D vector. Keypoints denote 2D keypoints comprising $18{\times}2$ body and $42{\times}2$ hand coordinates. Visual features are features extracted from the ViT backbone of TokenHMR~\cite{dwivedi2024tokenhmr}. All features are flattened and normalized across the dataset before being input to the encoder. }
\label{tab:feature_dims}
\end{table}

\FloatBarrier
\clearpage

\section{Win Ratios (TAG-Bench-v0, VBench-2.0)}
\label{sec:win_ratios}
We show the win ratios obtained in Sec.~\ref{sec:vbench} (VBench-2.0~\cite{vbench2}) and Sec.~\ref{sec:compare} (TAG-Bench-v0) below.

\begin{table}[h]
\centering
\small
\resizebox{0.6\textwidth}{!}{
\begin{tabular}{@{}lcc@{}}
\hline
\textbf{Model} & \textbf{Human win ratio} & \textbf{Metric win ratio} \\
\hline
Sora-480p~\cite{cho2024sora}      & 0.76 & 0.63 \\
Kling 1.6~\cite{kling2024}        & 0.78 & 0.83 \\
Hunyuan~\cite{kong2024hunyuanvideo} & 0.65 & 0.53 \\
CogVideoX-1.5~\cite{hong2022cogvideo} & 0.00 & 0.14 \\
\hline
\end{tabular}
}
\caption{\footnotesize \textbf{Win ratios for {\tempname} on the VBench-2.0 human-anatomy subset}: As detailed in Sec.~\ref{sec:vbench}, the ranking of models inferred by our {\tempname} ({\tempscore}) metric aligns with human judgment.}
\label{tab:vbench_anatomy}
\end{table}

\begin{table}[h]
\centering
\small
\resizebox{0.6\textwidth}{!}{
\begin{tabular}{@{}lcc@{}}
\hline
\textbf{Model} & \textbf{Human win ratio} & \textbf{Metric win ratio} \\
\hline
Hunyuan~\cite{kong2024hunyuanvideo}      & 0.39 & 0.40 \\
Opensora~\cite{openai2024sora}           & 0.23 & 0.30 \\
Runway Gen-4~\cite{runway2024gen4}       & 0.63 & 0.65 \\
Wan2.2~\cite{wan2025wan}                 & 0.90 & 0.77 \\
Wan2.1~\cite{wan2025wan}                 & 0.33 & 0.35 \\
\hline
\end{tabular}
}
\caption{\footnotesize \textbf{Win ratios for {\actionname} on TAG-Bench-v0}: As detailed in Sec.~\ref{sec:compare}, the ranking of models inferred by our {\actionname} ({\actionscore}) metric aligns with human judgment.}
\label{tab:tag_consistency}
\end{table}

\begin{table}[h]
\centering
\small
\resizebox{0.6\textwidth}{!}{
\begin{tabular}{@{}lcc@{}}
\hline
\textbf{Model} & \textbf{Human win ratio} & \textbf{Metric win ratio} \\
\hline
Hunyuan~\cite{kong2024hunyuanvideo}      & 0.42 & 0.48 \\
Opensora~\cite{openai2024sora}           & 0.17 & 0.28 \\
Runway Gen-4~\cite{runway2024gen4}       & 0.59 & 0.61 \\
Wan2.2~\cite{wan2025wan}                 & 0.91 & 0.72 \\
Wan2.1~\cite{wan2025wan}                 & 0.39 & 0.34 \\
\hline
\end{tabular}
}
\caption{\footnotesize \textbf{Win ratios for {\tempname} on TAG-Bench-v0}: As detailed in Sec.~\ref{sec:compare}, the ranking of models inferred by our {\tempname} ({\tempscore}) metric aligns with human judgment.}
\label{tab:vbench_temporal}
\end{table}

\newpage

\section{Additional Experiments}
\label{sec:addl_experiments}
\begin{figure}[h]
  \centering
  \includegraphics[width=0.45\linewidth]{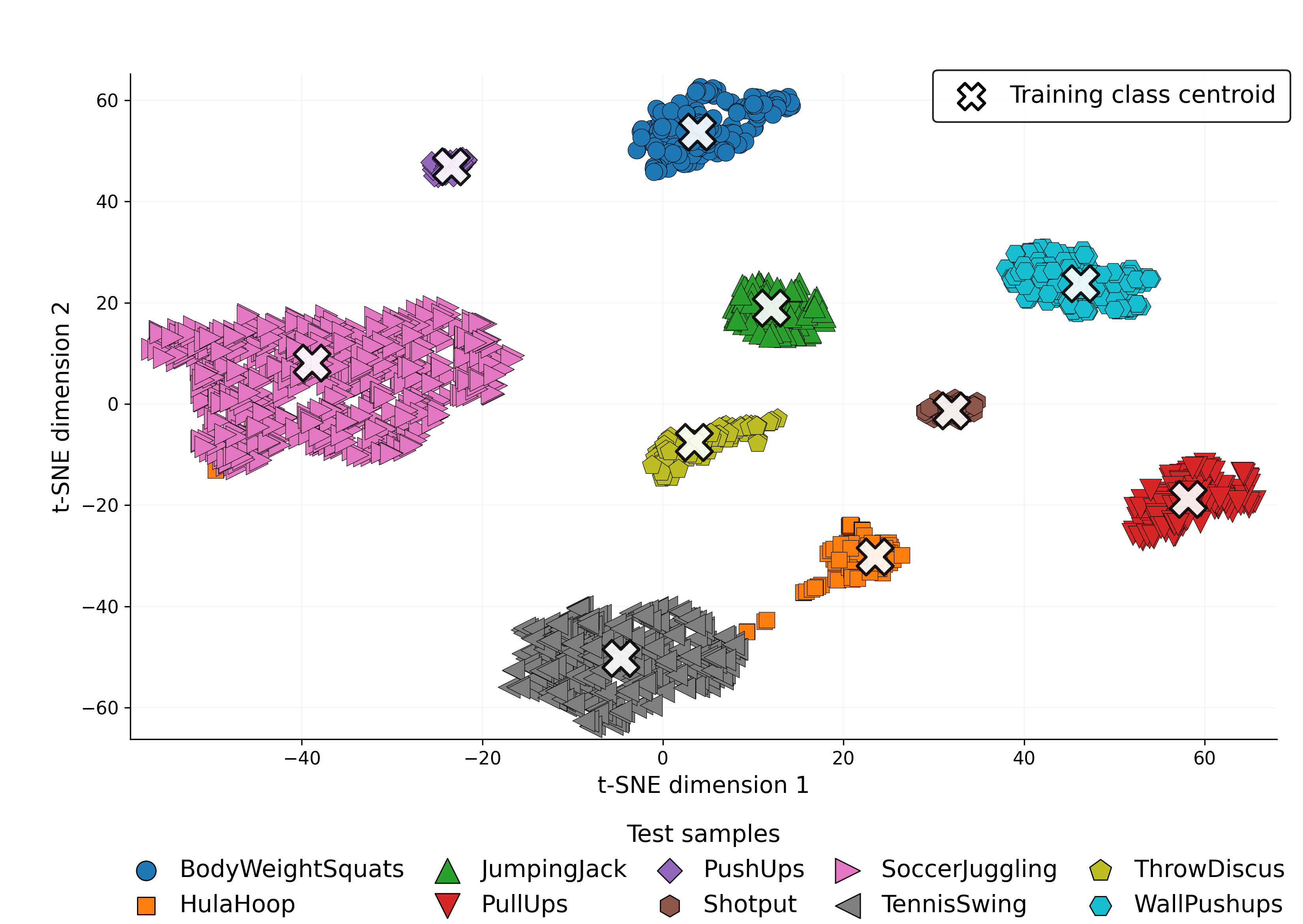}
  \caption{\footnotesize \textbf{t-SNE visualization of the embeddings of unseen test videos of diverse actions.} We project the $z_{\text{CLS}}$ embeddings of \emph{unseen real} test videos (colored markers) and the corresponding \emph{training} class centroids (crosses) using t-SNE~\cite{tsne}. It is evident that unseen test videos cluster around their respective class centroids, indicating that the learned embedding space captures compact and semantically meaningful action structure.
  }
  \label{fig:tsne_clusters_real}
\end{figure}

\subsection{Action class separability}
\label{subsec:act_class_sep}
We evaluate whether the learned embedding space places unseen real videos into the correct action-specific regions. For this, we extract the temporal window-level embeddings ($z_{\text{CLS}}$) (Sec.~\ref{sec:embedding_model}) from the held-out test set, run $K$-means clustering on these embeddings with $K{=}10$ (matching the number of action classes (Sec.~\ref{sec:human_eval})), and measure alignment of the cluster assignments with the ground-truth labels using Normalized Mutual Information (NMI)~\cite{nmi}. 
We obtain a high NMI score of $0.98$, indicating that unseen videos map to the correct action-specific region in the embedding space.
Figure~\ref{fig:tsne_clusters_real} visualizes this separation using t-SNE~\cite{tsne}: we plot embeddings of test windows along with the corresponding class centroids computed from the training set. Each action forms a tight, well-separated cluster around its real-video centroid. This shows that the embedding space captures a generalizable notion of action semantics, and that class centroids derived from real videos serve as reliable reference points for evaluating generated videos.\\

\newpage


\begin{figure}[h]
  \centering
  \begin{subfigure}[b]{0.4\linewidth}
    \centering
    \includegraphics[width=\linewidth]{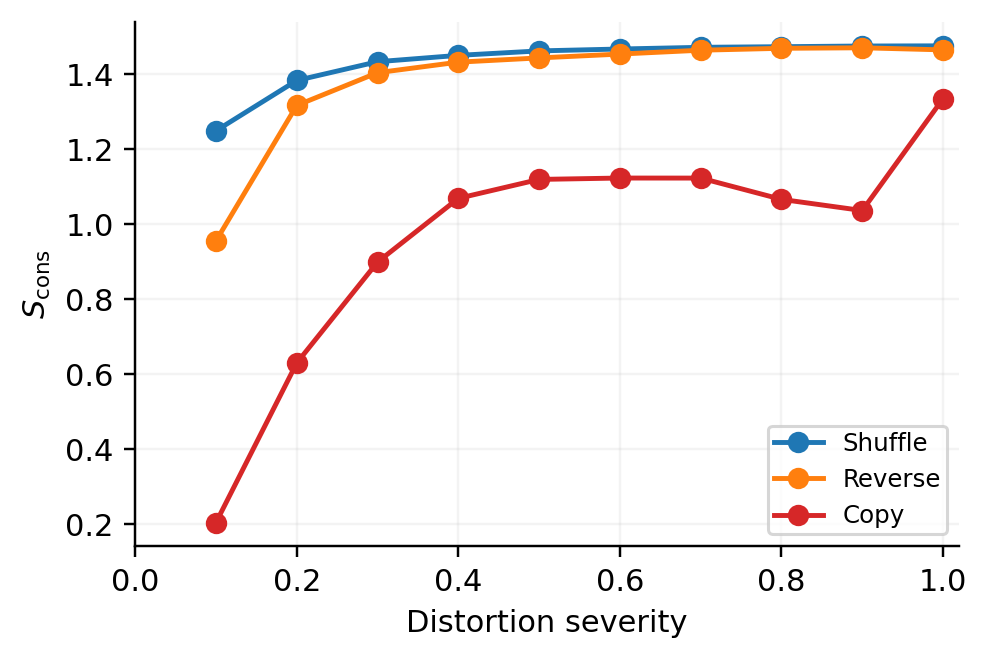}
  \end{subfigure}%
  \hspace{0.05\linewidth} 
  \begin{subfigure}[b]{0.4\linewidth}
    \centering
    \includegraphics[width=\linewidth]{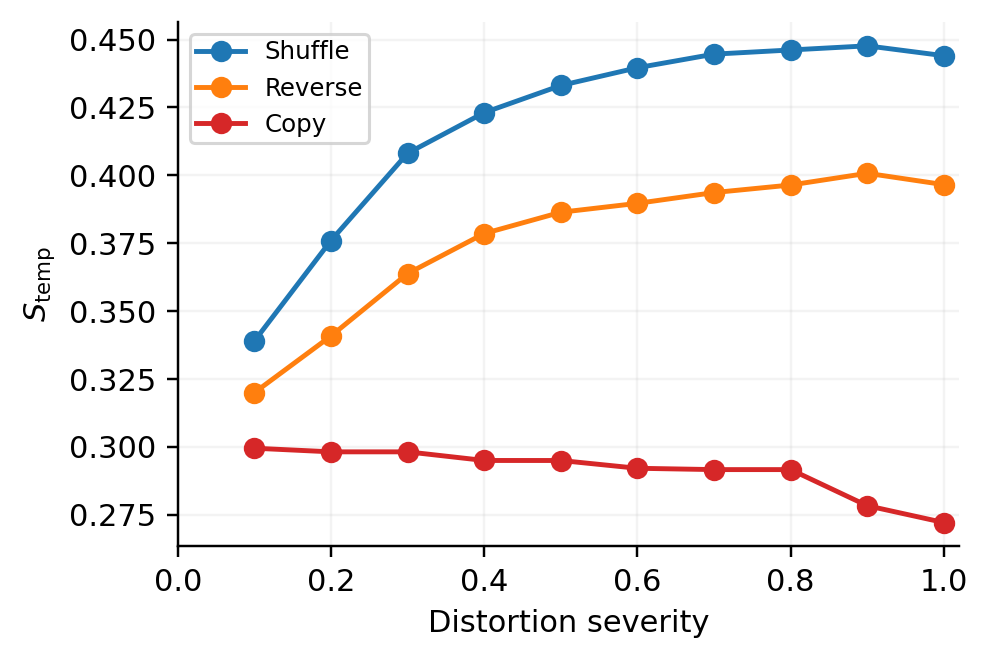}
  \end{subfigure}

    \caption{\footnotesize {Sensitivity of {\actionname} ({\actionscore}) and {\tempname} ({\tempscore}) to controlled temporal distortions. The mean value across all test samples is plotted for each distortion type and severity.}}
  \label{fig:distortion_sensitivity}
\end{figure}

\subsection{Sensitivity to temporal distortions} 
\label{subsec:temp_dist}
We measure whether the learned embedding space and metrics are temporally sensitive. For this, we apply controlled temporal corruptions to unseen real videos as done on training data (Sec.~\ref{sec:objective}). We describe each perturbation below:

\begin{itemize}
  \item \textbf{Shuffle}: randomly reorders a portion of frames, breaking local motion continuity. Severity controls the number of shuffled frames: 0.1 shuffles 10\% of frames, while 1.0 shuffles all frames in the window.
  
  \item \textbf{Reverse}: reverses the frame order in a temporal segment of the window, inverting the temporal direction of motion. Severity determines the length of the reversed segment: at 0.1, only 10\% of frames are reversed, while 1.0 reverses the entire window.
  
  \item \textbf{Copy}: replaces a contiguous segment of frames with copies of the first frame. Severity specifies how much of the sequence is replaced: 0.1 replaces 10\% of the frames, whereas 1.0 replaces all frames with the first frame of the window.
\end{itemize}

We then compute both {\actionscore} and {\tempscore} on these perturbed videos.
As shown in Fig.~\ref{fig:distortion_sensitivity}, both metrics monotonically increases (where lower is better, as described in Sec.~\ref{sec:embedding_metrics}) with increasing distortion severity: {\actionscore} increases as corrupted videos drift away from the real-action manifold, and {\tempscore} increases as temporal smoothness deteriorates (with the exception of ``Copy", due to identical adjacent frames resulting in low differences in their frame embeddings (Sec.~\ref{sec:embedding_metrics})). This confirms that the learned space is explicitly sensitive to temporal dynamics and does not rely solely on visual appearance at the frame level.\\

\newpage

\subsection{Effect of temporal window length}
\label{subsec:window_exp}
We train separate models using different temporal window sizes ($T\in\{4,8,16,32,64,128,256\}$). 
For videos shorter than $T$ frames, the final frame is repeated to match the window length. As shown in Table~\ref{tab:temporal_window}, performance improves significantly when increasing $T$ from $4$ to $32$ ($0.39~{\rightarrow}~\textcolor{ForestGreen}{0.61})$, for {\actionname}), indicating the importance of sufficient temporal context. However, increasing $T$ beyond $32$ yields no further gains while substantially increasing computational cost. Thus, we adopt $T{=}32$ for all experiments.\\
\begin{table}[H]
\centering
\resizebox{0.55\textwidth}{!}{%
\begin{tabular}{lcccccccc}
\toprule
\textbf{Window size ($T$)} & 4 & 8 & 16 & 32 & 64 & 128 & 256 \\
\midrule
{Action Consistency} & 0.39 & 0.46 & 0.59 & \textbf{0.61} & 0.56 & 0.57 & 0.55  \\
{Temporal Coherence} & 0.43 & 0.43 & 0.60 & \textbf{0.64} & 0.58 & 0.61 & 0.59 \\
\bottomrule
\end{tabular}}
\caption{\footnotesize \textbf{Effect of temporal window size.} Increasing the temporal window length $T$ improves performance significantly up to $T{=}32$, beyond which no gains are observed, while computational cost rises sharply.}
\label{tab:temporal_window}
\end{table}

\newpage

\subsection{Visualizing Outliers From Fig.~\ref{fig:tsne_clusters}}
\label{subsec:outlier_exp}

\begin{figure}[!h]
  \centering
  \includegraphics[width=0.99\linewidth, keepaspectratio]{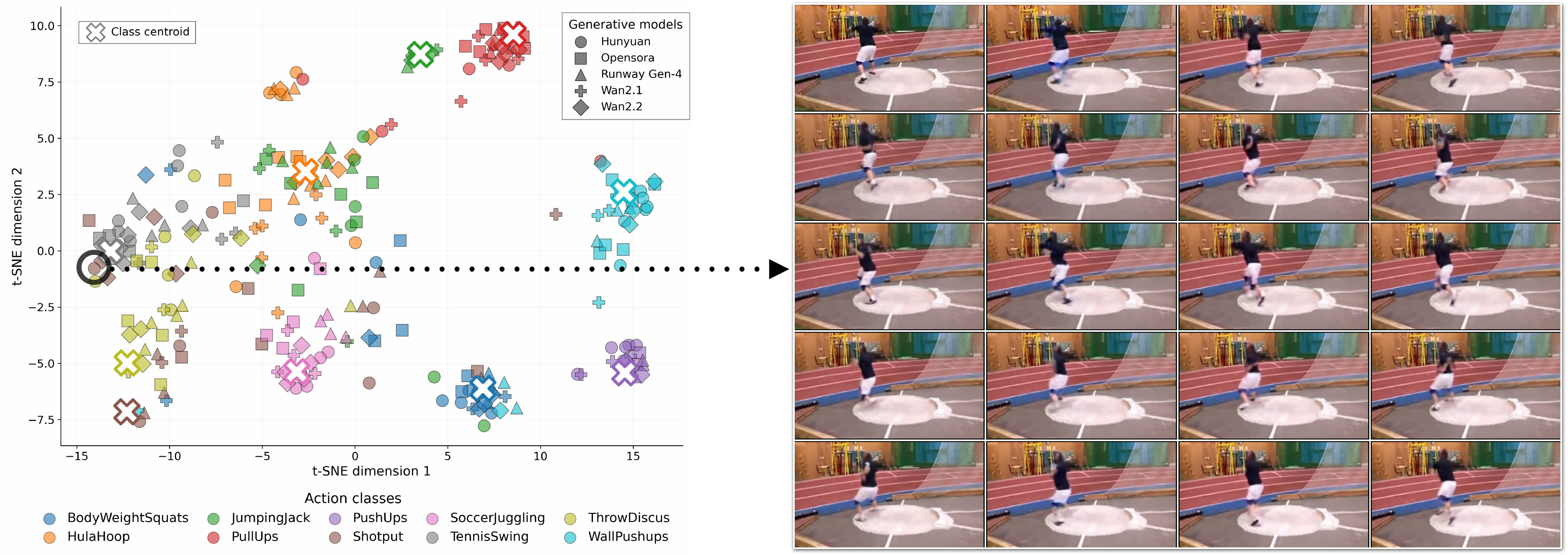}
  \caption{\footnotesize \textbf{Example failure case for the Shotput class.} 
The highlighted point corresponds to a generated video from Hunyuan intended to depict \textit{Shotput}, whose motion is temporally inconsistent and exhibits noticeable anatomical artifacts, as shown by the uniformly sampled frames on the right. 
Due to these irregularities, the resulting embedding does not clearly align with any specific action cluster in the feature space, reflecting the incoherent motion dynamics of the generated video.}
  \label{fig:outlier_shotput}
\end{figure}

\begin{figure}[!h]
  \centering
  \includegraphics[width=0.99\linewidth, keepaspectratio]{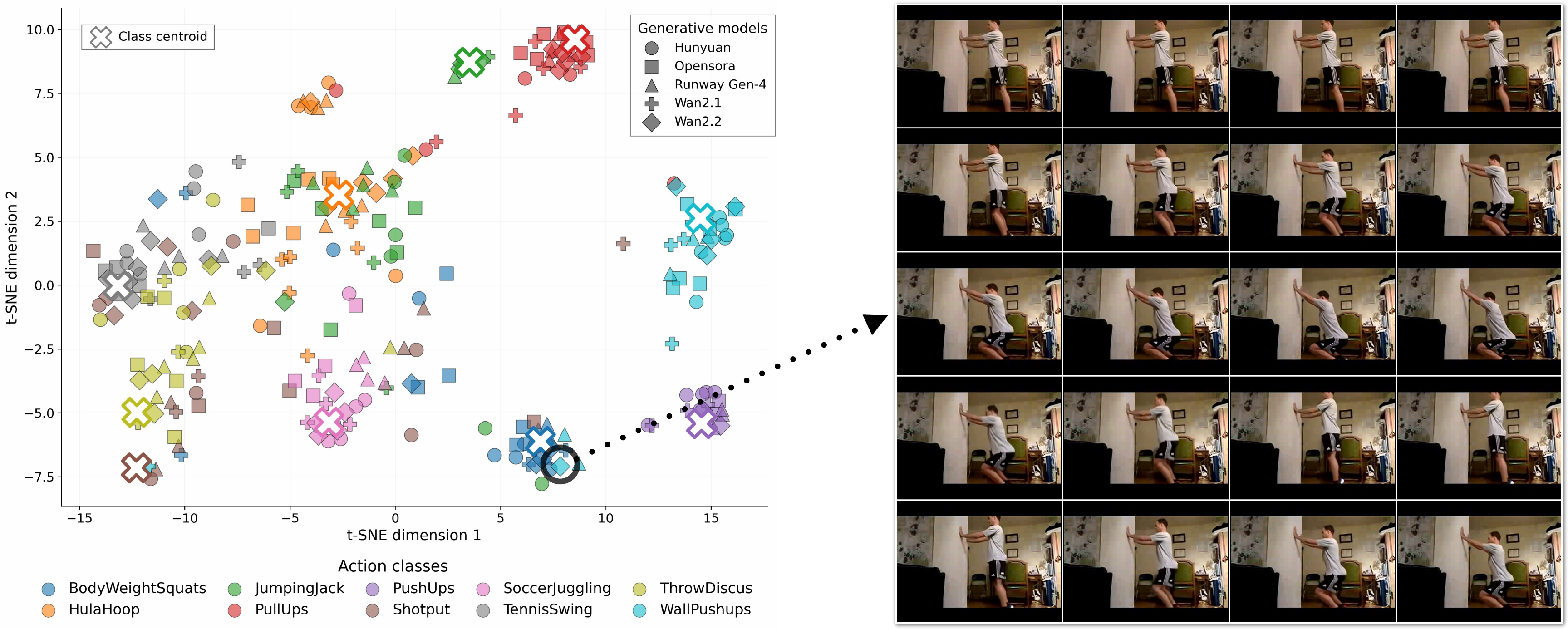}
\caption{\footnotesize \textbf{Example failure case for the WallPushups class.} 
The highlighted point corresponds to a generated video from Wan2.2 intended to depict \textit{WallPushups}, but the generated motion resembles squats (as shown by the uniformly sampled frames on the right). 
Consequently, its representation lies closer to the \textit{BodyWeightSquats} cluster and far from the \textit{WallPushups} cluster, indicating that the model misinterpreted the action and produced motion inconsistent with the intended class.}
  \label{fig:outlier_wallpushups}
\end{figure}

\subsection{Does providing detailed prompt help generate better actions?} 
\label{subsec:llm_exp}
To study this, we use Runway-Gen4~\cite{runway2024gen4} to generate video pairs from $30$ UCF101 initial frames all belonging to the complex action of ``soccer juggling.'' Specifically, we compare videos generated from the original short prompt (Sec.~\ref{sec:tagbench}) against those from LLM-enhanced prompts. We report the win ratio (Sec.~\ref{sec:vbench}) based on our AC and TC metrics, defined as the fraction of pairs in which the enhanced-prompt video receives a better score. 
The enhanced prompts win in 55\% of cases; this marginal improvement suggests that richer descriptions alone do not reliably improve action or motion realism.

\subsection{Generated videos as hard-negatives} 
\label{subsec:gen_neg_exp}
In addition to temporally distorted real videos (Sec.~\ref{sec:objective}), we explore using generated videos from CogVideoX, which are notably poor at depicting the intended action and motion, as additional hard-negatives during training. This yields only marginal changes in performance ({\actionname}: $0.61~{\rightarrow}~\textcolor{ForestGreen}{0.62}$, {\tempname}: $0.64~{\rightarrow}~\textcolor{Red}{0.63}$), suggesting that our temporally distorted real-video negatives already provide a strong training signal.

\subsection{More training data yields better correlation} 
\label{subsec:train_exp}
We test whether additional training data improves the learned embedding space by augmenting UCF101~\cite{soomro2012ucf101} with Kinetics-700~\cite{k700} clips depicting the same action classes. As Kinetics videos are not temporally trimmed to only showcase the action~\cite{k700}, we adopt an active sampling approach: our UCF101-trained model (Sec.~\ref{sec:implementation_details}) scores each temporal window from Kinetics-700 by its distance to the corresponding class centroid. By retaining only the most representative 20\% windows (i.e., with least distances) and augmenting the training set with these samples, correlation scores improve from $0.61~{\rightarrow}~\textcolor{ForestGreen}{0.65}$ for {\actionname} and $0.64~{\rightarrow}~\textcolor{ForestGreen}{0.65}$ for {\tempname}. However, performance drops when incorporating samples beyond the top 20\% (i.e., less representative windows farther from the class centroid). For example, using the top 30\% reduces {\actionname} from $0.65~{\rightarrow}~\textcolor{Red}{0.63}$, and reduces {\tempname} from $0.65~{\rightarrow}~\textcolor{Red}{0.63}$.

\FloatBarrier
\clearpage

\subsection{Attention weights}
\label{subsec:attentn_weights}
\begin{figure}[h]
  \centering
  \includegraphics[width=0.4\linewidth, keepaspectratio]{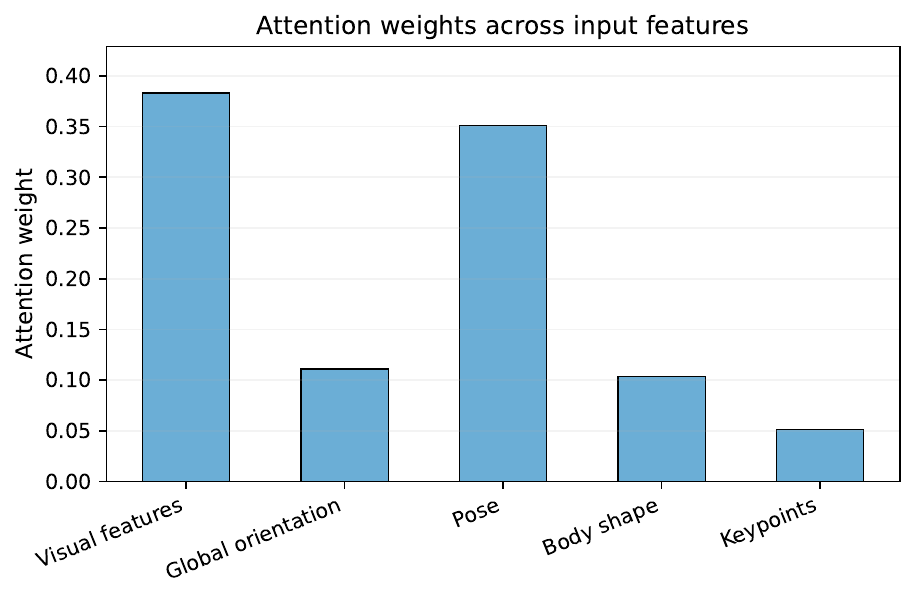}
  \caption{\footnotesize Average attention weights averaged over all real test videos.}
  \label{fig:attn1}
\end{figure}

\begin{figure}[h]
  \centering
  \includegraphics[width=\linewidth, keepaspectratio]{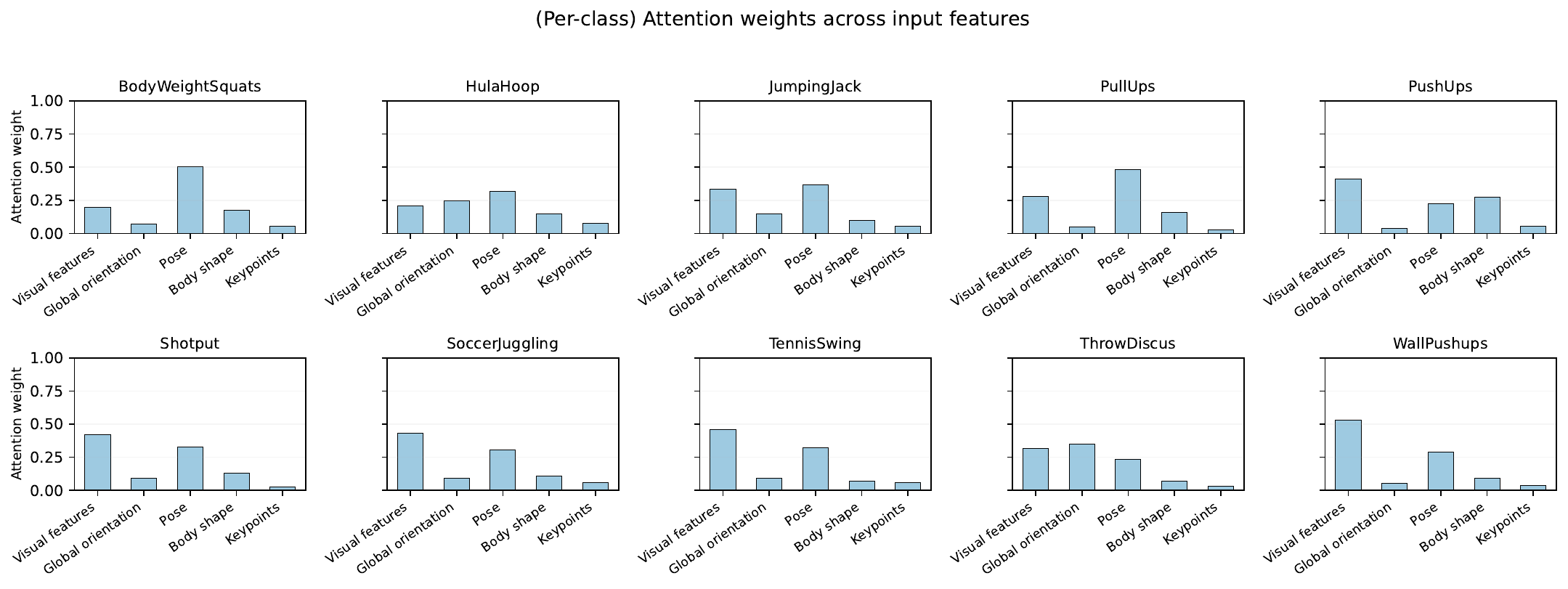}
  \caption{\footnotesize Per-class attention weights, averaged over all real test videos.}
  \label{fig:attn2}
\end{figure}

We visualize the attention weights from our embedding model (Sec.~\ref{sec:embedding_model}) in Fig.~\ref{fig:attn1} and Fig.~\ref{fig:attn2}. 
Figure~\ref{fig:attn1} shows the average attention distribution over all real test samples, while Figure~\ref{fig:attn2} breaks this down by action class.  We observe that attention weights vary by action class. Although the visual features (ViT features from TokenHMR~\cite{dwivedi2024tokenhmr}) dominate overall, these features inherently encode both appearance cues and implicit geometric structure (see Fig.~\ref{fig:tokenhmr}), as they are used to infer the SMPL parameters (Sec.~\ref{sec:3d_feats}). The model also assigns high weight to 3D pose features, indicating that anatomically grounded signals are essential for modeling human action. For certain actions, the model adapts its focus toward other input features, for example, the ``HulaHoop'' class shows increased reliance on global orientation.

\newpage

\subsection{Design choices}
\label{subsec:arch_design}

To handle heterogeneous features (e.g., 3D pose $\theta \in \mathbb{R}^{207}$, ViT features $f_{vis} \in \mathbb{R}^{1024}$), as in GENMO~\cite{li2025genmo}, we first project them to a common dimension and fuse them per-frame before temporal modeling with a Transformer. Let $\mathbf{E}_t$ denote the set of feature embeddings at frame $t$. We compared four fusion strategies to obtain the frame-level feature $\mathbf{f}_t$: \textbf{(a) No fusion}: Concatenating $\mathbf{E}_t$ and deferring fusion to the Transformer's self-attention (AC=$0.53$, TC=$0.51$), \textbf{(b) 1D convolution}: Stacking features in $\mathbf{E}_t$ and applying a depth-wise 1D convolution (AC= $0.55$, TC=$0.58$)
\textbf{(c) Uniform sum}: Element-wise summation of $\mathbf{E}_t$ (AC=0.59, TC=0.62), \textbf{(d) Attention-weighted sum (ours)}: a learnable query vector to attend to the features in $\mathbf{E}_t$ via scaled dot-product attention, producing scalar weights for each feature, whose weighted sum gives $\mathbf{f}_t$, obtaining AC=0.61, TC=0.64.
The drop in (a) shows that explicit intra-frame fusion helps before temporal aggregation.
Our attention-weighted approach (d) excels as it captures action-specific importance; e.g., notice that the model prioritizes global orientation for \textit{HulaHoop} but 3D pose for \textit{PullUps} (Fig.~\ref{fig:attn2}). 

\subsection{Sensitivity to number of samples}
\label{subsec:num_samples}

We re-trained our model with varying number of samples per class, keeping rest of the setup fixed (Sec.~5.1). While our full training set inherits the long-tail distribution of UCF101 (ranging from $42$ to $137$ samples per class after filtering (L435)), our framework is also remarkably data-efficient. 
Table~\ref{rebuttal_tab1} shows that performance stabilizes with only $30$ samples per class. \emph{ With just $10$ samples, our model still achieves an AC of $0.55$, exceeding GPT-5’s $0.45$}, demonstrating robustness to limited and imbalanced training data.

\begin{table}[!h]
\centering
\scriptsize
\label{tab:data_efficiency}
\begin{tabular}{l|ccccccc}
\hline
No. of samples/class & 3 & 5 & 10 & 30 & 50 & 100 & All \\
\hline
AC & 0.43 & 0.49 & 0.55 & 0.61 & 0.60 & 0.58 & 0.61 \\
TC & 0.27 & 0.39 & 0.54 & 0.59 & 0.58 & 0.64 & 0.64 \\
\hline
\end{tabular}
\caption{Number of training samples per class v/s AC and TC.}
\label{rebuttal_tab1}
\end{table}

\FloatBarrier
\clearpage

\section{Baselines}

\begin{table}[t]
\small
\centering
\resizebox{0.6\linewidth}{!}{
\begin{tabular}{l|cc}
\toprule
\textbf{Method} &
\begin{tabular}{c}
\textbf{Corr. with Action} \\
\textbf{Consistency} \textcolor{ForestGreen}{$\uparrow$}
\end{tabular} &
\begin{tabular}{c}
\textbf{Corr. with Temporal} \\
\textbf{Coherence} \textcolor{ForestGreen}{$\uparrow$}
\end{tabular} \\
\midrule
Random & -0.07 & -0.11 \\
\midrule
\multicolumn{3}{c}{\textbf{Feature-based automatic metrics}} \\
\midrule
PIQUE~\cite{PIQUE}                        & -0.19 & -0.13 \\
BRISQUE \cite{BRISQUE}       & -0.04 & 0.01 \\
CLIP-sim \cite{clip}                    & 0.03 & 0.16 \\
DINO-sim \cite{dino}                     & 0.08 & 0.21 \\
SSIM-sim \cite{ssim}                     & -0.08 & -0.04 \\
MSE-dyn~\cite{1284395}                      & -0.15 & -0.08 \\
SSIM-dyn~\cite{1284395}                     & -0.06 & -0.03 \\
CLIP-Score \cite{clipscore}                  & 0.08 & 0.00 \\
X-CLIP-Score \cite{ma2022xclipendtoendmultigrainedcontrastive}                & 0.00 & -0.07 \\
TRAJAN \cite{trajan}         & -0.12 & -0.12 \\
VideoMAE(UCF101)-classification        & 0.18 & 0.17 \\
VBench-2.0 \cite{vbench2} (Human Anatomy)   & -0.40 & 0.02 \\
VBench-2.0 \cite{vbench2} (Human Identity)      & 0.06 & 0.02 \\
VBench-2.0 \cite{vbench2} (Human Clothes)   & 0.12 & 0.11 \\
\midrule
\multicolumn{3}{c}{\textbf{MLLM-based fine-tuned metrics}} \\
\midrule
\faUnlock\; VideoScore \cite{videoscore} (Visual Quality)  & -0.12 & -0.06 \\
\faUnlock\; VideoScore \cite{videoscore} (Temporal Consistency)  & -0.09 & -0.04 \\
\faUnlock\; VideoScore \cite{videoscore} (Dynamic Degree)  & -0.19 & -0.16 \\
\faUnlock\; VideoScore \cite{videoscore} (T2V Alignment)      & -0.07 & -0.04 \\
\faUnlock\; VideoScore \cite{videoscore} (Factual Consistency)   & -0.14 & -0.08 \\
\faUnlock\; VideoScore2 \cite{videoscore2} (Visual Quality) & 0.14 & 0.16 \\
\faUnlock\; VideoScore2 \cite{videoscore2} (T2V Alignment)     & 0.17 & 0.09 \\
\faUnlock\; VideoScore2 \cite{videoscore2} (Physical Consistency) & 0.18 & 0.17 \\
\faUnlock\; VideoPhy-2 \cite{videophy2} (Semantic Adherence)       & 0.19 & 0.16 \\
\faUnlock\; VideoPhy-2 \cite{videophy2} (Physical Commonsense)     & 0.28 & 0.37 \\
\midrule
\multicolumn{3}{c}{\textbf{MLLM Prompting}} \\
\midrule
\faUnlock\; LLaVA-1.5-7B \cite{llava1p5}               & -0.17 & -0.14 \\
\faUnlock\; LLaVA-v1.6-mistral-7b-hf \cite{llava1p6}    & -0.10 & 0.18 \\
\faUnlock\; Idefics2-8B \cite{idefics2}                 & -0.05 & -0.06 \\
\faUnlock\; Qwen3-VL-8B-Instruct \cite{qwen3}        & 0.34 & 0.28 \\
\faLock\; Gemini-2.5-Flash \cite{gemini2p5}            & 0.40 & 0.25 \\
\faLock\; Gemini-2.5-Pro \cite{gemini2p5}              & 0.31 & 0.26 \\
\faLock\; GPT-4o \cite{gpt4o}                      & 0.34 & 0.31 \\
\faLock\; GPT-5 \cite{gpt5}                       & \underline{0.45} & \underline{0.38} \\
\midrule
\multicolumn{3}{c}{\textbf{Ours}} \\
\midrule
{\actionname} {\actionscore} (Ours)    & \textbf{0.61} & 0.45 \\
{\tempname} {\tempscore} (Ours)          & 0.53 & \textbf{0.64} \\
\midrule
$\Delta$ over best baseline & + 0.16 & + 0.26 \\
Relative improvement (\%) over best baseline & + 35.6\% & + 68.4\% \\
\midrule
\multicolumn{3}{c}{\emph{Inter-rater agreement}} \\
\midrule
Human vs Human & 0.72 & 0.71 \\
\bottomrule
\end{tabular}
}
\caption{\textbf{Correlation (Spearman's $\rho$) between model predictions and human scores for {\actionname} and {\tempname}.} (Higher is better). 
`VideoMAE(UCF101)-classification' uses the confidence score ~\cite{tong2022videomae} as the predicted scores. \faUnlock\; denotes open-source models, while \faLock\; denotes closed-source models. We observe that the proposed {\actionscore} outperforms all methods for {\actionname}, and  {\tempscore} for {\tempname}. The next best performing metric is \underline{underlined}.}
\label{tab:baselines_suppl}
\end{table}

\label{sec:baseline_F}
We describe the baselines we report in Table~\ref{tab:baselines} below.

\subsection{Feature-based Metrics}
\label{subsec:baseline_f1}
\noindent\textbf{TRAJAN~\cite{trajan}.} Evaluates motion realism by reconstructing point trajectories using a video autoencoder; inconsistent or implausible motion yields lower Average Jaccard scores (0–1), with higher scores indicating more natural and realistic videos.

\noindent\textbf{PIQUE~\cite{PIQUE}.}
A no-reference image quality metric that estimates distortion by analyzing local blocks in each frame and scoring only perceptually significant regions; outputs a 0–100 distortion score, where lower values indicate better visual quality.
 
\noindent\textbf{BRISQUE~\cite{BRISQUE}.} A no-reference quality metric that detects deviations from natural image statistics on a per-frame basis, producing a 0–100 distortion score (lower is better).

\noindent\textbf{SSIM similarity~\cite{ssim}.} Computes frame-wise structural similarity between generated and reference videos and averages scores across frames (0–1, higher is better).

\noindent\textbf{CLIP Similarity.} Computes the average cosine similarity of CLIP VIT-B/32~\cite{clip} embeddings for adjacent frames.

\noindent\textbf{DINO Similarity.} Computes the average cosine similarity of DINO ViT-B/16~\cite{caron2021emergingpropertiesselfsupervisedvision} embeddings for adjacent frames.

\noindent\textbf{MSE Dyn.~\cite{1284395}.} Computes the average mean squared error of every fourth frame.

\noindent\textbf{SSIM~\cite{1284395} Dyn.} Computes the structural similarity of every fourth frame based on luminance, contrast, and spatial arrangement. 

\noindent\textbf{CLIP Score~\cite{clipscore}} Computes the average cosine similarity of the CLIP ViT-B/32~\cite{clip} embedding of a given text prompt and the CLIP embeddings of each frame. Each text prompt follows the format ``A person doing [action]."

\noindent\textbf{X-CLIP Score.} Computes the cosine similarity of X-CLIP~\cite{ma2022xclipendtoendmultigrainedcontrastive} embeddings of a given text prompt and X-CLIP video embeddings. The text prompts each follow the format ``A person doing [action]."

\noindent\textbf{VBench-2.0~\cite{vbench2} (Human Anatomy).} Evaluates frame-level anatomical correctness by detecting abnormalities in body structure, hands, and faces using three anomaly-detection models; the final score is the percentage of frames without detected anomalies.

\noindent\textbf{VBench-2.0~\cite{vbench2} (Human Face).} Measures whether the same person is preserved across frames by computing facial feature similarity (ArcFace~\cite{deng2019arcface}) relative to the first frame, ignoring segments with multiple or no people.

\subsection{Multi-modal Large Language Models (MLLMs) based metrics}
\label{subsec:baseline_f2}

\noindent\textbf{VBench-2.0~\cite{vbench2} (Human Clothes).} Assesses whether the person's clothing remains consistent across the video by probing an MLLM (LLaVA-video-7B~\cite{zhang2024video}).

\noindent\textbf{Videophy2~\cite{videophy2}} Measures the physical commonsense and semantic adherence of a video. Physical commonsense captures if physical rules (e.g. gravity, collision dynamics) are obeyed. Semantic adherence measures the adherence of a video to its text description. 



\noindent\textbf{Videoscore~\cite{videoscore}.}  
Evaluates generated videos across five dimensions: visual quality, temporal consistency, dynamic degree, text–video alignment, and factual consistency, using a model trained on large-scale human ratings; each dimension is scored from 1 to 4.

\noindent\textbf{Videoscore2~\cite{videoscore2}.}  
Evaluates videos along visual quality, text–video alignment, and physical consistency using a model trained on large-scale human ratings; outputs scores from 1 to 5 along with reasoning for each dimension.

\section{MLLM Prompting}
\label{sec:mllm_eval}

\begin{figure}[H]
    \centering
    \includegraphics[width=\linewidth]{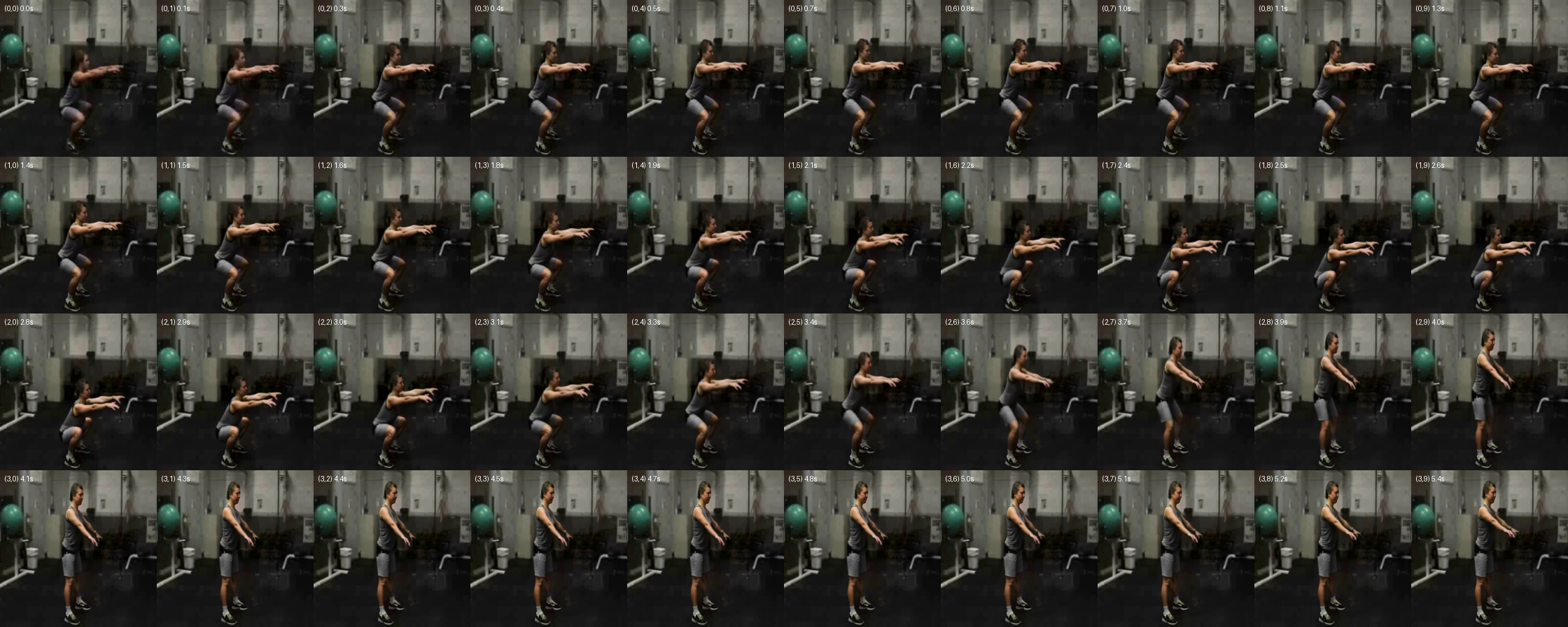}
    \caption{Example of the $4{\times}10$ grid-panel layout used to prompt MLLMs. 
    Shown here is a video generated by \textbf{Hunyuan}~\cite{kong2024hunyuanvideo} for the action class \textit{BodyWeightSquats}. We uniformly sample 40 frames and place them in row-major order. Each cell overlays its grid coordinates \texttt{(row,col)} in the top-left; when the video duration is available, the timestamp is also shown (e.g., \texttt{(0,3) 1.2s}). This grid preserves temporal progression and spatial structure, providing clearer visual evidence than direct video input.}
    \label{fig:human_eval_ui}
\end{figure}

\subsection{Prompt used for MLLM evaluation}
\label{subsec:mllm_prompt}
We provide the exact instruction shown to MLLMs for all panel-based evaluations.

{\small
\begin{Verbatim}[breaklines=true]
You are an expert evaluator of AI-generated video quality.

Your job is to analyze a grid of frames extracted from ~5 seconds of video
and score TWO axes:

1. Action Consistency (action_consistency):
   How well does the visible action in the frames match the described target action?
   - Focus only on what is clearly shown.
   - Check pose, motion pattern, timing, and repeated evidence of that action.
   - Do NOT guess intentions outside the frames.

2. Temporal Coherence (temporal_coherence):
   How physically realistic / plausible are the motions and body configurations?
   - Look for broken limbs, impossible joint angles, teleporting limbs,
     limbs merging into objects, obvious gravity violations,
     ghost artifacts (extra arms / missing torso), etc.
   - Minor render glitches are OK if motion is still basically human-plausible.
   - Very warped anatomy or impossible motion should score low.

You must base your judgment ONLY on what is visible in the panel.
Treat the panel as a time grid read left-to-right, top-to-bottom.

Return JSON ONLY with:
- "action_consistency": float in [0,1]
- "temporal_coherence": float in [0,1]
- "confidence": float in [0,1]  (your confidence in these scores)
- "evidence": a list of AT MOST 3 items. Each item is { "cell": {"row": int, 
"col": int}, "description": "short reason" }. Choose the 1-3 most diagnostic cells.
- "rationale": one or two concise sentences summarizing both axes

Scoring guide for BOTH action_consistency and temporal_coherence:
0.90-1.00: very strong evidence, consistent across many frames
0.70-0.89: mostly good, only small issues
0.40-0.69: mixed; frequent issues or uncertainty
0.10-0.39: mostly wrong / implausible / inconsistent
0.00-0.09: completely wrong, impossible, or not supported

Important:
- Use 0-indexed coordinates (row, col) when citing evidence.
- Do NOT invent frames you cannot see.
\end{Verbatim}
}

\subsection{Limited impact of in-context learning.}
\label{subsec:mllm_icl}
We also investigate whether in-context learning (ICL) can better align MLLM-based scores with human judgments, and in particular whether the outputs of our method can serve as useful demonstrations. We begin with a controlled case study on \textit{JumpingJack}, where we provide the model with a small set of good and bad examples together with short rationales. 
In this setting, ICL does not yield consistent improvements: relative to the baseline prompt, it slightly decreases correlation with human judgments on {\actionname} while only modestly improving correlation on {\tempname}. Moreover, the predicted scores remain heavily saturated near the extremes, suggesting that the model still tends to make binary-like decisions rather than nuanced motion-quality assessments.

\begin{table}[h]
    \centering
    \resizebox{\linewidth}{!}{
    \begin{tabular}{lcc}
        \toprule
        \textbf{Method} & \textbf{Corr. with Action Consistency} & \textbf{Corr. with Temporal Coherence} \\
        \midrule
        Qwen3-VL (baseline prompt)      & 0.34 & 0.28 \\
        Qwen3-VL (in-context learning, JumpingJack demos) & 0.29 & 0.32 \\
        \bottomrule
    \end{tabular}
    }
    \caption{Correlation (Spearman's $\rho$) between model predictions and human scores for {\actionname} and {\tempname} on the 300-video evaluation set.}
    \label{tab:qwen3_icl}
\end{table}

Table~\ref{tab:qwen3_icl} reports Spearman's $\rho$ between Qwen3's predictions and human scores. 
Relative to the baseline prompt, in-context learning \emph{reduces} the correlation for {\actionname} from $0.34$ to $0.29$, while it \emph{increases} the correlation for {\tempname} from $0.28$ to $0.32$. 
Thus, this in-context learning scheme introduces a trade-off between the two dimensions rather than yielding consistent gains, highlighting the need for structured, human-centric evaluation strategies—such as our learned action manifold.



\begin{table}[h]
\centering
\begin{tabular}{l @{\hspace{1.5em}} c @{\hspace{2em}} c}
\toprule
\textbf{Model} & \textbf{Baseline} & \textbf{In-context learning} \\
\midrule
GPT-5 & 0.45 & \textbf{0.47} \\
Gemini-2.5-Pro & 0.31 & \textbf{0.34} \\
\bottomrule
\end{tabular}
\caption{Spearman's $\rho$ between MLLM predictions and human {\actionname} scores across 10 action classes. In-context learning uses demonstrations selected using our metric.}
\label{tab:icl_10class}
\end{table}

We also test whether in-context learning can better align MLLM's scores with human judgements.
To examine whether the outputs of our model can be used as in-context learning samples, we expanded the dataset to include 10 classes: \textit{BodyWeightSquats, HulaHoop, JumpingJack, PullUps, PushUps, Shotput, SoccerJuggling, TennisSwing, ThrowDiscus,} and \textit{WallPushups}. As the MLLMs, we used GPT-5~\cite{gpt5} and Gemini-2.5-Pro~\cite{gemini2p5}.
For the good examples, we randomly selected two real videos from UCF-101~\cite{soomro2012ucf101}. For the bad examples, we selected two videos generated by CogVideoX~\cite{yang2024cogvideox} with the highest AC scores.
As shown in Table~\ref{tab:icl_10class}, in this expanded setting GPT-5 achieves an {\actionname} correlation of 0.47 and Gemini-2.5-Pro achieves 0.34, outperforming their respective baselines. This suggests that our scores can indeed be used as informative demonstrations.

However, the observed gains are highly sensitive to the specific examples used, and performance varies substantially across action classes. Taken together, these results indicate that ICL may provide modest benefits, but its effects are neither robust nor consistently transferable across actions. This further motivates the need for more structured, human-centric evaluation strategies, such as our learned action manifold, rather than relying on prompting alone.

\newpage

\end{document}